\documentclass{article}

\usepackage{microtype}
\usepackage{graphicx}
\usepackage{subfigure}
\usepackage{booktabs} 

\usepackage{hyperref}

\usepackage[accepted]{icml2020}

\usepackage[T1]{fontenc}
\usepackage{mymath}

\icmltitlerunning{On the Noisy Gradient Descent that Generalizes as SGD}

\begin{document}

\twocolumn[
\icmltitle{On the Noisy Gradient Descent that Generalizes as SGD}



\icmlsetsymbol{equal}{*}

\begin{icmlauthorlist}
\icmlauthor{Jingfeng Wu}{jhu}
\icmlauthor{Wenqing Hu}{must}
\icmlauthor{Haoyi Xiong}{baidu}
\icmlauthor{Jun Huan}{ai}
\icmlauthor{Vladimir Braverman}{jhu}
\icmlauthor{Zhanxing Zhu}{pku}
\end{icmlauthorlist}

\icmlaffiliation{jhu}{Johns Hopkins University, Baltimore, MD, USA}
\icmlaffiliation{must}{Missouri University of Science and Technology, Rolla, MO, USA}
\icmlaffiliation{baidu}{Big Data Laboratory, Baidu Research, Beijing, China}
\icmlaffiliation{ai}{Styling.AI Inc., Beijing, China}
\icmlaffiliation{pku}{Peking University, Beijing, China}

\icmlcorrespondingauthor{Jingfeng Wu}{uuujf@jhu.edu}
\icmlcorrespondingauthor{Zhanxing Zhu}{zhanxing.zhu@pku.edu.cn}

\icmlkeywords{SGD, Noise, Regularization}

\vskip 0.3in
]



\printAffiliationsAndNotice{}  

\begin{abstract}
The gradient noise of SGD is considered to play a central role in the observed strong generalization abilities of deep learning. While past studies confirm that the magnitude and  covariance structure of gradient noise are critical for regularization, it remains unclear whether or not the class of noise distributions is important. In this work we provide negative results by showing that noises in classes different from the SGD noise can also effectively regularize gradient descent. Our finding is based on a novel observation on the structure of the SGD noise: it is the multiplication of the gradient matrix and a sampling noise that arises from the mini-batch sampling procedure. Moreover, the sampling noises unify two kinds of gradient regularizing noises that belong to the Gaussian class: the one using (scaled) Fisher as covariance and the one using the gradient covariance of SGD as covariance. Finally, thanks to the flexibility of choosing noise class, an algorithm is proposed to perform noisy gradient descent that generalizes well, the variant of which even benefits large batch SGD training without hurting generalization.
\end{abstract}

\section{Introduction}

Stochastic gradient descent (SGD) is one of the standard workhorses for optimizing deep models~\cite{bottou1991stochastic}.
Though initially proposed to remedy the computational bottleneck of gradient descent (GD), recent studies suggest SGD in addition induces a crucial implicit regularization, which prevents the over-parameterized models from converging to the minima that cannot generalize well~\cite{zhang2017understanding,zhu2018anisotropic,jastrzkebski2017three,hoffer2017,keskar2016large}.
To gain intuitions, one can compare the generalization abilities of 
(i) GD vs. SGD, (ii) small batch SGD vs. large batch SGD, and (iii) SGD vs. gradient Langevin dynamic (GLD).
Empirical studies confirm that 
(i) SGD outperforms GD~\cite{zhu2018anisotropic}, 
(ii) small batch SGD generalizes better than large batch SGD~\cite{hoffer2017,keskar2016large},
and (iii) GLD cannot compete with SGD~\cite{zhu2018anisotropic}.
To understand why these phenomena happen, let us look at the differences between the compared algorithms.
Firstly SGD can be viewed as GD, an deterministic algorithm, with an unbiased noise inserted at every iteration, which is called the \emph{gradient noise}~\cite{bottou2018optimization}.
Secondly the gradient noise of the small batch SGD has a much larger magnitude than that of the large batch SGD~\cite{hoffer2017,jastrzkebski2017three}.
Thirdly, even though the noise magnitude is tuned to be equal, the SGD noise has a nontrivial covariance structure, instead of just being a white noise as in GLD~\cite{zhu2018anisotropic}.
The above discussions exhibit a critical fact: 

\emph{Certain noises can effectively regularize gradient descent}.

Despite the efforts spent,
this important yet implicit regularization effect induced by noise has never been fully understood.
From the Bayesian perspective, the noise is interpreted to perform variational inference~\cite{mandt2017stochastic,chaudhari2017stochastic}.
Such interpretation, however, requires unrealistic assumptions such as the noise has constant covariance~\cite{mandt2017stochastic} or certain force is conservative~\cite{chaudhari2017stochastic}.
Another theory argues that the noise enables the gradient algorithm to escape from sharp minima~\cite{zhu2018anisotropic,hu2019quasi,simsekli2019tail} that typically generalize worse~\cite{hochreiter1997flat,keskar2016large}.
Hence GD enhanced by such noise tends to find flat minima that generalize well.
This explanation hold valid to some extent;
but the escaping behavior is too subtle to fit practice --- the loss/accuracy does not jump significantly after the dynamic reaching a minimum, e.g., see the final epochs of Figure~4 in~\cite{huang2017densely}. 
Therefore the algorithm does not explicitly escape from minima in practice.
Although the mechanism has not been completely understood, we can still recognize and utilize such implicit regularization by studying the properties of gradient noise.

We next summarize three important aspects of gradient noise that might introduce the regularization effects: noise magnitude, covariance structure and distribution class of noise. 

\textbf{Noise magnitude}~
The large batch SGD encounters performance deterioration compared with the small batch one, thus the magnitude of gradient noise matters~\cite{hoffer2017,keskar2016large,smith2018bayesian}.
Furthermore, \citet{jastrzkebski2017three} show that the ratio of learning rate to batch size, which directly controls the noise magnitude, has an important influence on the generalization of SGD: in a certain range, greater the ratio, larger the noise, and better the generalization.

\textbf{Noise covariance structure}~
From the perspective of escaping from minima, \citet{zhu2018anisotropic} emphasize the importance of the noise covariance structure for regularization.
They show that when the noise covariance contains curvature information, it performs better for escaping from sharp minima~\cite{zhu2018anisotropic,hu2019quasi,daneshmand2018escaping}.
Surprisingly, the covariance of the SGD noise aligns with the Hessian of the loss surface to some extent~\cite{zhu2018anisotropic,li2019hessian}, which then partly explains the benefits brought by the SGD noise.

\textbf{Noise class}~
Many works assume that the SGD noise belongs to the Gaussian class due to the classical central limit theorem ~\cite{ahn2012bayesian,chen2014stochastic,shang2015covariance,mandt2017stochastic,zhu2018anisotropic}.
Nonetheless, \citet{simsekli2019tail} first argue that the second moment of SGD noise might not exist, thus the Gaussianity assumption requires a second thought, since the classical central limit theorem has to be revised for heavy-tailed distributions~\cite{[Gnedenko-Kolmogorov],[Bertoin]}.
Instead in this case, the central limit theorem leads to Levy distribution which they adopt for modeling SGD noise.
By assuming so they obtain a faster escaping behavior of SGD~\cite{simsekli2019tail,nguyen2019first,csimcsekli2019heavy}.
Later \citet{panigrahi2019non} directly perform Gasussianity testing during the process of SGD learning deep neural networks.
They empirically find that when the batch size is greater than $256$, the SGD noise can be treated as Gaussian in the early phase of training;
but in general the SGD noise does not have to be Gaussian alike.

While past studies confirm the importance of noise magnitude and covariance structure, \emph{the role of noise class in regularizing a gradient method has not been fully explored}.
In this work, we attempt to address this issue from a novel perspective of sampling noise.
Taking SGD for instance, we notice the gradient noise is indeed caused by the mini-batch sampling procedure. This observation enables us to establish a key notion called the \emph{sampling noise} to characterize the stochasticity of mini-batch sampling.
Based on the sampling noise, we show that noises in classes different from the SGD noise can also effectively regularize gradient descent, thus provide negative evidence on the impact of the noise class.
On the other hand, thanks to the flexibility of choosing noise class, we are allowed to use noisy gradient descent with best fitted noises based on practical requirements, beyond the vanilla SGD.
This finding supports the methods to employ structured Gaussian noises for improving GD/large batch SGD~\cite{zhu2018anisotropic,wen2019interplay}.

\textbf{Contributions}~
In summary we obtain the following important results:
\begin{enumerate}
\item 
A novel perspective is proposed for interpreting the SGD noise: it is the multiplication of a gradient matrix and a \emph{sampling noise} which raises from the mini-batch sampling process. 
A general class of noisy gradient descent is thus defined based on the sampling noise.
\item 
The regularization role of the distribution class of gradient noise is then investigated.
In both theory and experiments, we demonstrate that the noise class might not be a crux for regularization, provided suitable noise magnitude and covariance structure.
\item 
Two kinds of gradient regularizing noises from the Gaussian classes are then revised, i.e., the one using the (scaled) Fisher as covariance~\cite{wen2019interplay} and the one employing the gradient covariance of SGD as covariance~\cite{zhu2018anisotropic}.
The equivalence between them is established by analyzing their sampling noises.
\item 
Thanks to the unimportance of the noise class, 
an algorithm is proposed to perform generalizable noisy gradient descent with noises from various classes.
Its variant even benefits large batch SGD training without hurting generalization.
\end{enumerate}

\section{The gradient noise of SGD}\label{sec:sgd-noise}
Let the training data be $\{x_i\}_{i=1}^n$, and consider the empirical loss $L(\theta) = \frac{1}{n}\sum_{i=1}^n \ell(x_i;\theta)$, where $\ell(x;\theta)$ is the loss over one sample and $\theta \in \Rbb^d$ is the parameter to be optimized.
Define the \emph{loss vector} as
$\Lcal (\theta) = \left( \ell(x_1;\theta), \dots, \ell(x_n;\theta) \right) \in \Rbb^{1\times n}$, then the \emph{gradient matrix} is $\grad_{\theta} \Lcal (\theta) =\left( \grad_{\theta}\ell(x_1;\theta), \dots, \grad_{\theta}\ell(x_n;\theta) \right)\in \Rbb^{d \times n}$.
Let $\mathbbm{1} = (1,\dots,1)^T \in \Rbb^{n}$,
then $L(\theta) =\frac{1}{n}\Lcal(\theta)\cdot\mathbbm{1}$.

\textbf{SGD}~
During each iteration of SGD, the algorithm first randomly draws a mini-batch of samples with index set $B_t = \{i_{1}, \dots, i_{b}\}$ in size $|B_t| = b$, and then performs parameter update using the \emph{stochastic gradient} $\tilde{g}(\theta)$ computed by the mini-batch and learning rate $\eta$,
\begin{equation*}
    \theta_{t+1} = \theta_t -\eta \tilde{g}(\theta_t),\quad 
    \tilde{g}(\theta_t) = \frac{1}{b}\sum_{i \in B_t} \grad_{\theta} \ell(x_i; \theta_t).
\end{equation*}

\textbf{Sampling noise}~
Note that the stochasticity of $\tilde{g}(\theta_t)$ is caused by the randomness of the mini-batch sampling procedure, thus the stochastic gradient could be written as
\begin{equation*}
    \tilde{g}(\theta_t) = \grad_{\theta} \Lcal(\theta_t) \cdot \Wcal_{\sgd},
\end{equation*}
where $\Wcal_{\sgd} \in \Rbb^n$ is a random \emph{sampling vector} characterizing the mini-batch sampling process.
For instance considering mini-batch SGD without replacement, the sampling vector $\Wcal_{\sgd}$ contains exactly $b$ multiples of $\frac{1}{b}$ and $n-b$ multiples of zero with random index.
It is easy to see that $\Ebb [\Wcal_{\sgd}] = \frac{1}{n}\mathbbm{1}$, thus $\Ebb[\tilde{g}(\theta_t)]  =\frac{1}{n}\grad_{\theta}\Lcal(\theta_t)\cdot\mathbbm{1} = \grad_{\theta}L(\theta_t)$, i.e., the stochastic gradient $\tilde{g}(\theta_t)$ is an unbiased estimator of the full gradient $\grad_{\theta}L(\theta_t)$.

Define the \emph{sampling noise} as $\Vcal_{\sgd} = \Wcal_{\sgd} - \frac{1}{n}\mathbbm{1}$.
Then the stochastic gradient has the decomposition of
\begin{equation*}
    \tilde{g}(\theta_t) = \grad_{\theta}L(\theta_t) + \grad_{\theta} \Lcal(\theta_t) \cdot \Vcal_{\sgd},\quad \Ebb [\Vcal_{\sgd}] = 0.
\end{equation*}
The first two moments of $\Vcal_{\sgd}$ are given in Proposition~\ref{thm:sampling-noise}.

\begin{prop}(Mean and covariance of the SGD sampling noise)\label{thm:sampling-noise}
For mini-batch sampled without replacement, the SGD sampling noise $\Vcal_{\sgd}$ satisfies
\begin{equation*}
    \Ebb [\Vcal_{\sgd}] = 0,\quad  
    \Var [\Vcal_{\sgd}] = \frac{n-b}{bn(n-1)} \left( I- \frac{1}{n}\mathbbm{1}\mathbbm{1}^T \right).
\end{equation*}
For mini-batch sampled with replacement, the SGD sampling noise $\Vcal_{\sgd}^{\prime}$ satisfies
\begin{equation*}
    \Ebb [\Vcal_{\sgd}^{\prime}] = 0,\quad 
    \Var [\Vcal_{\sgd}^{\prime}] = \frac{1}{bn} \left( I- \frac{1}{n}\mathbbm{1}\mathbbm{1}^T \right).
\end{equation*}
\end{prop}
The proof is left in Section~\ref{sec:pf-sampling-noise} of the Supplementary Materials.
If not stated otherwise, we focus on SGD with replacement in the remaining parts.
However, our arguments hold for both of them with mild modifications.

\textbf{Gradient noise}~
From the viewpoint of sampling noise, the \emph{gradient noise} of SGD is the multiplication of the gradient matrix and its sampling noise,
\begin{equation*}
    \upsilon_{\sgd} (\theta_t) = \tilde{g}(\theta_t) - \grad_{\theta}L(\theta_t) = \grad_{\theta}\Lcal(\theta_t) \cdot \Vcal_{\sgd}.
\end{equation*}
Note that while the sampling noise $\Vcal_{\sgd}$ is state-independent, the gradient noise $\upsilon_{\sgd} (\theta_t)$ is coupled with the parameter $\theta_t$.
By Proposition~\ref{thm:sampling-noise}, the first two moments of the gradient noise are $\Ebb[\upsilon_{\sgd}(\theta_t)] = \grad_{\theta} \Lcal (\theta_t) \Ebb[\Vcal_{\sgd}] = 0$ and
\begin{equation}\label{eq:sgd-cov}
\begin{aligned}
    &C(\theta_t) = \Var [\upsilon_{\sgd}(\theta_t)] = \grad_{\theta} \Lcal (\theta_t) \Var[\Vcal_{\sgd}] \grad_{\theta} \Lcal (\theta_t)^T \\
    &= \frac{1}{b}\left(\frac{1}{n}\grad\Lcal(\theta_t)\grad\Lcal(\theta_t)^T - \grad L(\theta_t) \grad L(\theta_t)^T\right) .
\end{aligned}
\end{equation}
In the following we call $C(\theta_t)$ the \emph{SGD covariance}.

As the structure of the SGD noise is clear, we turn to discuss the properties of the noise that affect its implicit regularization.
Studies on large batch SGD training~\cite{keskar2016large,hoffer2017} exhibit the importance of the \emph{noise magnitude}, which is controlled by $\sqrt{\tfrac{\eta}{b}}$~\cite{jastrzkebski2017three}.
And from the viewpoint of escaping from minima, the implicit bias of SGD is also closely related to the \emph{noise covariance structure} $C(\theta)$
~\cite{zhu2018anisotropic,hu2019quasi,li2019hessian}.
Recently, the role of the \emph{noise class} raises research interests, as discussed below.

\subsection{The class of the SGD noise}
Due to the i.i.d. sampling of a mini-batch, as the batch size approaches infinity, the theory about  limit theorems guarantees that the SGD noise converges to certain infinite divisible distribution~\cite{[Gnedenko-Kolmogorov], [Bertoin]}.
If the second moment of the noise is finite, the limiting infinite divisible distribution will belong to the Gaussian class. 
Thus many works assume the Gaussianity of the SGD noise~\cite{chen2014stochastic,ahn2012bayesian,shang2015covariance,mandt2017stochastic,jastrzkebski2017three,zhu2018anisotropic}.
However, if the second moment does not exist, so that the noise is heavy-tailed, then the gradient noise should converge to a Levy type distribution, as assumed by~\cite{simsekli2019tail,nguyen2019first,csimcsekli2019heavy}.
Moreover, it is also questionable whether in practice the batch size is large enough for applying limit theorems.
We investigate the two issues in the following.

\textbf{The finiteness of the SGD covariance}~
Based on analysis of the structure of the SGD noise, we have $C(\theta_t) = \grad_{\theta} \Lcal (\theta_t) \Var[\Vcal_{\sgd}] \grad_{\theta} \Lcal (\theta_t)^T$ by Eq.~\eqref{eq:sgd-cov},
and $\Var[ \Vcal_{\sgd}]$ is finite by Proposition~\ref{thm:sampling-noise}.
Thus if the gradient matrix $\grad_{\theta} \Lcal (\theta_t)$ is bounded (almost everywhere), then $C (\theta_t)$ must be finite (almost everywhere).
Firstly, the typical components of neural networks are twice differentiable (almost everywhere)~\cite{goodfellow2016deep};
moreover, with common deep learning tricks such as near-zero initialization, early stopping, learning rate decay, weight decay, etc, the optimization process only happens in a small area around the near-zero initialization~\cite{neyshabur2017,jacot2018neural,cao2019generalization}.
Therefore it is reasonable to assume that the gradient matrix $\grad_{\theta} \Lcal (\theta_t)$ is bounded almost everywhere in the area of our concerns.
Thereby we argue that it is safe to assume the finiteness of the SGD covariance.

\textbf{The non-Gaussianity of the SGD noise}~
Even with finite covariance, it is still unclear whether in practice the batch size is sufficiently large for the Gaussian to be a good approximation for the SGD noise, especially when it comes to the extremely high dimensional parameter in deep learning.
To validate this, \citet{panigrahi2019non} directly perform Gaussianity tests to the SGD noise during the training of deep neural networks. 
They empirically find that when the batch size is greater than $256$, the SGD noise behaves like a Gaussian one in the early phase of training; but generally the SGD noise does not belong to the Gaussian class.

\textbf{The impact of the noise class}~
We conclude that the SGD noise belongs to a particular distribution class that is neither Levy nor Gaussian.
One might wonder if this particular distribution class of SGD noise is crucial for its regularization effects.
In the remaining of this work, we address this issue by studying a general framework of noisy gradient descent which can employ noises from various classes, including the SGD noise class and the Gaussian class.
The framework is called the \emph{multiplicative SGD} (MSGD).

\subsection{Multiplicative SGD}
During each iteration, the proposed MSGD randomly generates a sampling vector $\Wcal\in \Rbb^{n}$ with mean as $\Ebb[\Wcal] = \frac{1}{n}\mathbbm{1}$, and then takes update
\begin{equation*}
    \theta_{t+1} = \theta_{t} - \eta \grad_{\theta} \Lcal(\theta_t) \Wcal.
\end{equation*}
Denote the sampling noise as $\Vcal = \Wcal - \frac{1}{N}\mathbbm{1}$, then the gradient noise is $\upsilon(\theta_t) = \grad_{\theta} \Lcal(\theta_t) \Vcal$.
Since our goal is to study the impact of noise class, the covariance of the gradient noise thus has to be fixed for excluding the influences of the noise magnitude and covariance structure.
To this end it is sufficient to fix the covariance of the sampling noise, i.e., $\Var[\Vcal] = \Var [\Vcal_{\sgd}]$.
The MSGD can then be written as
\begin{equation}\label{eq:msgd}
\begin{aligned}
    & \theta_{t+1} = \theta_{t} - \eta \grad_{\theta} L(\theta_t) + \eta \grad_{\theta} \Lcal(\theta_t) \Vcal,\\ 
    &\text{where}\quad \Ebb[\Vcal] = 0,\quad \Var[\Vcal] = \Var [\Vcal_{\sgd}].
\end{aligned}
\end{equation}
In the MSGD iteration~\eqref{eq:msgd}, the gradient noise $\upsilon(\theta_t)$ is decided by the deterministic gradient matrix $\grad_{\theta} \Lcal(\theta_t)$ and a sampling noise $\Vcal$.
Thus we can control the class of the gradient noise by choosing the class of the sampling noise.
For example, the gradient noise $\upsilon(\theta_t)$ becomes the SGD noise if $\Vcal = \Vcal_{\sgd}$.
Besides, if the sampling noise belongs to the Gaussian class, i.e., $\Vcal_{\text{G}} \sim \Ncal \left(0, \Var[\Vcal_{\sgd}] \right)$, then the gradient noise $\upsilon_{\text{G}}(\theta_t)=\grad_{\theta}\Lcal(\theta_t)\Vcal_{\text{G}}$ is also Gaussian, i.e., $\upsilon_{\text{G}}(\theta_t) \sim \Ncal \left(0, C(\theta_t)\right)$, where $C(\theta_t) = \grad_{\theta} \Lcal (\theta_t) \Var[\Vcal_{\sgd}] \grad_{\theta} \Lcal (\theta_t)^T$ by Eq.~\eqref{eq:sgd-cov}.
In this case we call the iteration~\eqref{eq:msgd} the \emph{Gaussian MSGD}.
Moreover, gradient noises in other classes of practical interests can also be obtained with suitable sampling noises, e.g., Bernoulli sampling noises and sparse Gaussian sampling noises.

We then explore the role of the noise class by studying the generalization abilities of the MSGD iteration~\eqref{eq:msgd} with noises from different classes.

\section{Theoretical study}\label{sec:theory}
We first theoretically revise the role of the noise class for regularizing the algorithm.
For the solution $\hat{\theta}$ found by noisy gradient descent and the optimal parameter $\theta_*$, the generalization error can be measured as $\Ebb_{x, \hat{\theta}}\left[\ell(x; \hat{\theta}) - \ell(x; \theta_*)\right]$.
Now suppose the loss function $\ell(x; \theta)$ can be approximated by a quadratic one (with respect to $\theta$), then the generalization error involves just the first two moments of $\hat{\theta}$, which depends on at most the second moment information about the gradient noise, since the noise only accumulates linearly in the final solution $\hat{\theta}$ because of the linearity of the gradient.
Hence intuitively, provided the noise covariance, the generalization error has little dependence on the particular class that the gradient noise belongs to.

To formalize the above intuition, we follow the setting of~\cite{bach2013non,dieuleveut2017harder,defossez2015averaged} and consider an online linear regression problem 
\begin{equation}\label{eq:loss}\tag{$\Pcal$}
\min_{\theta} f(\theta) := \half \Ebb_{(x,y)}[(x^T \theta - y)^2].
\end{equation}
Let $\Sigma = \Ebb_{x} [x x^T]$, then $f(\theta)$ always admits an optimal $\theta_* = \Sigma^{\dagger} \Ebb_{(x,y)}[y x]$.
Denote the residual as $\epsilon = y - x^T \theta_*$, then $\Ebb[\epsilon x] = 0$.
We also adopt the following standard assumptions~\cite{bach2013non,dieuleveut2017harder,defossez2015averaged}:
\begin{align}
    & \Ebb \left[\norm{x}_2^2 x x^T\right] \preceq R^2 \Sigma; \tag{$\Acal_1$}\label{eq:asp-cov} \\
    & \Ebb\left[\epsilon^2 x x^T\right] \preceq \sigma^2 \Sigma; \tag{$\Acal_2$}\label{eq:asp-noise} \\
    & \Sigma \preceq \lambda I. \tag{$\Acal_3$}
\end{align}

\begin{rmk*}
The assumption~\eqref{eq:asp-cov} is satisfied when the data is almost surely bounded, i.e., $\norm{x}_2\le R$;
and~\eqref{eq:asp-noise} holds for almost surely bounded data or when the model is well-specified, i.e., $\epsilon_n$ is independent with $x_n$, and i.i.d. of zero mean and variance $\sigma^2$~\cite{dieuleveut2017harder}.
\end{rmk*}

Typically the problem~\eqref{eq:loss} is learned by the averaged solution $\bar{\theta}_n = \frac{1}{n+1}\sum_{i=0}^n\theta_i$ of the (small batch) SGD~\cite{bach2013non,dieuleveut2017harder,defossez2015averaged}
\begin{equation}\label{eq:online-sgd}
    \theta_{n+1} = \theta_n - \eta \sum_{r\in b_n} \left(x_r x_r^T \theta_n - y_r x_r \right),
\end{equation}
where $b_n$ is the index set of a randomly sampled mini-batch with a small batch size $|b_n| = b$.
We note $b$ could be $1$. 
To validate our understanding, 
we also consider the following (large batch) MSGD algorithm
\begin{equation}\label{eq:online-msgd}
    \theta_{n+1} = \theta_n - \eta \sum_{r\in B_n} w_r \left(x_r x_r^T \theta_n - y_r x_r \right),
\end{equation}
where $B_n$ is the index set of a randomly sampled mini-batch, with a relatively large batch size $|B_n| = B > b$, and $\Wcal = (w_{r_1}, \dots, w_{r_B})^T$ is a random sampling vector where $\Ebb[\Wcal] = \frac{1}{B}\mathbbm{1}$.

The following theorem characterizes the generalization error of the large batch MSGD~\eqref{eq:online-msgd} and the small batch SGD~\eqref{eq:online-sgd}.

\begin{thm}\label{thm:sgd-generalization}
Suppose the covariance of the sampling vector in MSGD~\eqref{eq:online-msgd} satisfies $\Var[\Wcal] = \frac{B-b}{bB(B-1)}\left(I - \frac{1}{B}\mathbbm{1}\mathbbm{1}^T\right)$.
Then for both of the large batch MSGD~\eqref{eq:online-msgd} and the small batch SGD~\eqref{eq:online-sgd}, we have
\begin{equation*}
    \Ebb_{\bar{\theta}_n}[f(\bar{\theta}_n)] - f(\theta_*) \le \frac{C_1}{n+1} + \frac{C_2}{(n+1)^2},
\end{equation*}
where $C_1$ and $C_2$ are constants that depend on $b$, $\eta$, $R$, $\sigma$, $\lambda$ and $\theta_0$, but not $B$.
\end{thm}
The proof is left in Supplementary Materials, Section~\ref{sec:pf-sgd-generalization}.
The generalization error bound is indeed optimal as it matches the statistical lower bounds in certain circumstances~\cite{dieuleveut2017harder}.

According to Theorem~\ref{thm:sgd-generalization},
provided appropriate noise covariance (see Proposition~\ref{thm:sampling-noise}),  the large batch MSGD generalizes as the small batch SGD, and its generalization does not depend on the specific class of its gradient noise.
Hence the noise class is not crucial for generalization, at least for the quadratic loss.
For general loss functions, we empirically validate our understanding in the next section.

\section{Empirical study}

In this section we present our empirical results.
The setup details are explained in Supplementary Materials, Section~\ref{sec:setups}.

To begin with, we propose Algorithm~\ref{alg:msgd} for efficiently performing the MSGD iteration~\eqref{eq:msgd}.
The key idea of Algorithm~\ref{alg:msgd} is that the gradient operator commutes with the multiplication operator.
Using Algorithm~\ref{alg:msgd}, we can easily inject noises with the SGD covariance to GD.

\begin{algorithm}
\caption{Multiplicative SGD}
\label{alg:msgd}
\begin{algorithmic}[1]
\STATE \textbf{Input}: Initial parameter $\theta_0\in \mathbb{R}^d$, training data $\{(x_i, y_i)\}_{i=1}^n$, loss function $\ell_i(\theta)=\ell((x_i, y_i), \theta)\in \Rbb$, loss vector $\Lcal(\theta) = \left(\ell_1(\theta),\dots,\ell_n(\theta)\right) \in \Rbb^{1\times n}$, learning rate $\eta>0$
\FOR {$k=0,1,2,...,K-1$}
    \STATE Generate a sampling noise $\Vcal\in \Rbb^{n}$ with zero mean and desired covariance
    \STATE Compute the sampling vector $\Wcal = \frac{1}{n}\mathbbm{1} + \Vcal$
    \STATE Compute the randomized loss $\tilde{L}(\theta_k) = \Lcal(\theta_k) \Wcal$
    \STATE Compute the stochastic gradient $\grad_{\theta} \tilde{L}(\theta_k)$
    \STATE Update the parameter $\theta_{k+1} = \theta_k-\eta \grad_{\theta}\tilde{L}(\theta_k)$
\ENDFOR
\STATE \textbf{Output}: Output $\theta_K$
\end{algorithmic}
\end{algorithm}

\subsection{Gaussian noise with SGD covariance}\label{sec:gen-gauss-noise}

In this part we discuss the ways to generate Gaussian gradient noises with covariance as the SGD covariance.
Such noises in the Gaussian class is of great importance for both theoretical analysis of the implicit regularization~\cite{zhu2018anisotropic,jastrzkebski2017three} and empirical algorithms for large batch SGD training~\cite{wen2019interplay}.
We denote the desired Gaussian noise as $\upsilon_{\text{G}}(\theta) \sim \Ncal\left(0,  C(\theta)\right)$, where $C(\theta)$ is the SGD covariance as defined in Eq.~\eqref{eq:sgd-cov}.

\textbf{SVD}~
The typical approach of generating $\upsilon_{\text{G}}(\theta)$ is based on the singular value decomposition (SVD)~\cite{zhu2018anisotropic}: 
one first computes the covariance matrix and then applies SVD on it, $C(\theta) = U(\theta) \Lambda(\theta) U(\theta)^T$, then transforms a white noise $\epsilon\in \Rbb^{d}$ into the Gaussian noise desired, $\upsilon_{\text{G}}(\theta) = U(\theta) \Lambda(\theta)^{\half} \epsilon$.

However, there are two obstacles in the above approach:
(i) evaluating and storing the covariance matrix $C(\theta)\in\Rbb^{d\times d}$ is computationally unacceptable, with both $n$ and $d$ being large;
(ii) performing SVD for a $d\times d$ matrix is comprehensively hard when $d$ is extremely large, e.g., deep neural networks.
Furthermore, (i) and (ii) repeat at every iteration of parameter update, since $C(\theta)$ depends on the parameter $\theta$.
In compromise, current works suggest to approximate $C(\theta)$ using only its diagonal or block diagonal elements~\cite{wen2019interplay,zhu2018anisotropic,jastrzkebski2017three,martens2015optimizing}.
Generally, there is no guarantee that the diagonal information could approximate the full SGD covariance well;
specifically, \citet{zhu2018anisotropic} demonstrate that such diagonal approximation cannot recover the regularization effects of SGD.
Thus a more effective approach of generating Gaussian noise with the SGD covariance is demanded. 

\textbf{Gaussian sampling noise}~
As discussed before, a gradient noise belongs to the Gaussian class if and only if its sampling noise is also Gaussian.
Thus based on the MSGD framework~\eqref{eq:msgd}, to insert a Gaussian gradient noise $\upsilon_{\text{G}}(\theta) \sim \Ncal\left(0,  C(\theta)\right)$, we only need to apply Algorithm~\ref{alg:msgd} with its corresponding Gaussian sampling noise, which is $\Vcal_{\text{G}} \sim \Ncal \left(0, \Var[\Vcal_{\sgd}]\right)$ according to Eq.~\eqref{eq:sgd-cov}.
Notice that the covariance of the SGD sampling noise admits a natural decomposition as
$\Var[\Vcal_{\sgd}] 
    = \frac{1}{bn} \left( I- \frac{1}{n}\mathbbm{1}\mathbbm{1}^T \right)
    = \frac{1}{bn} \left(I - \frac{1}{n}\mathbbm{1}\mathbbm{1}^T\right)  \left(I - \frac{1}{n}\mathbbm{1}\mathbbm{1}^T \right)^T$.
Thus the Gaussian sampling noise could be obtained by letting
$\Vcal_{\text{G}} = \frac{1}{\sqrt{bn}}\left(I - \frac{1}{n}\mathbbm{1}\mathbbm{1}^T\right) \epsilon$, where $\epsilon\in \Rbb^{n}$ is a white noise.
We use \emph{MSGD-Cov} to name this approach of injecting Gaussian gradient noise with the SGD covariance.

\begin{figure*}
\centering
\begin{tabular}{ccc}
\includegraphics[width=0.3\linewidth]{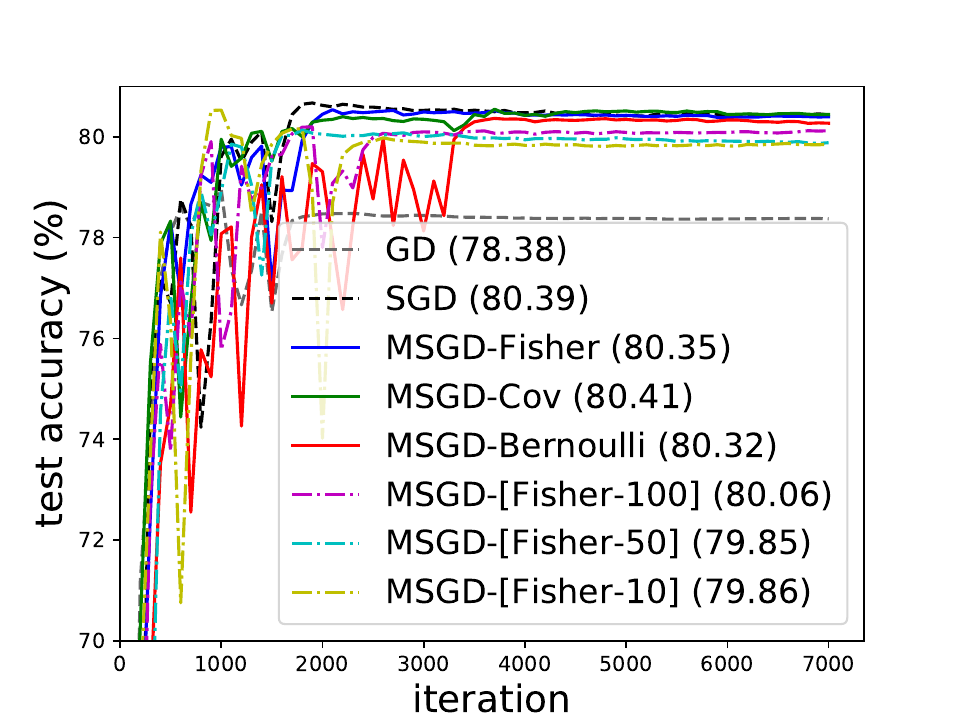} &
\includegraphics[width=0.3\linewidth]{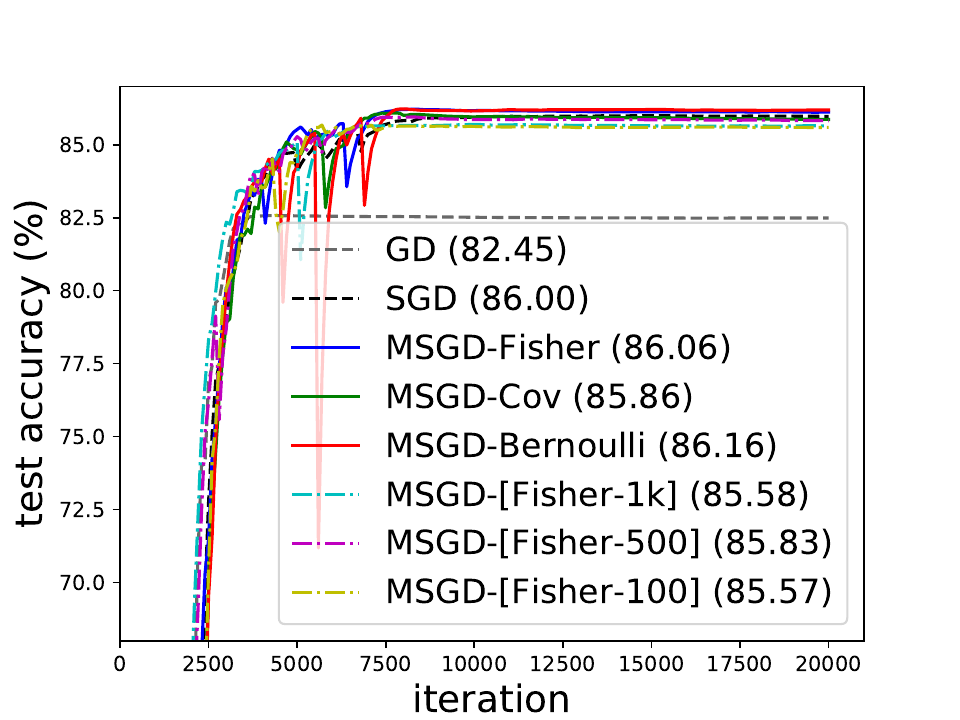} &
\includegraphics[width=0.3\linewidth]{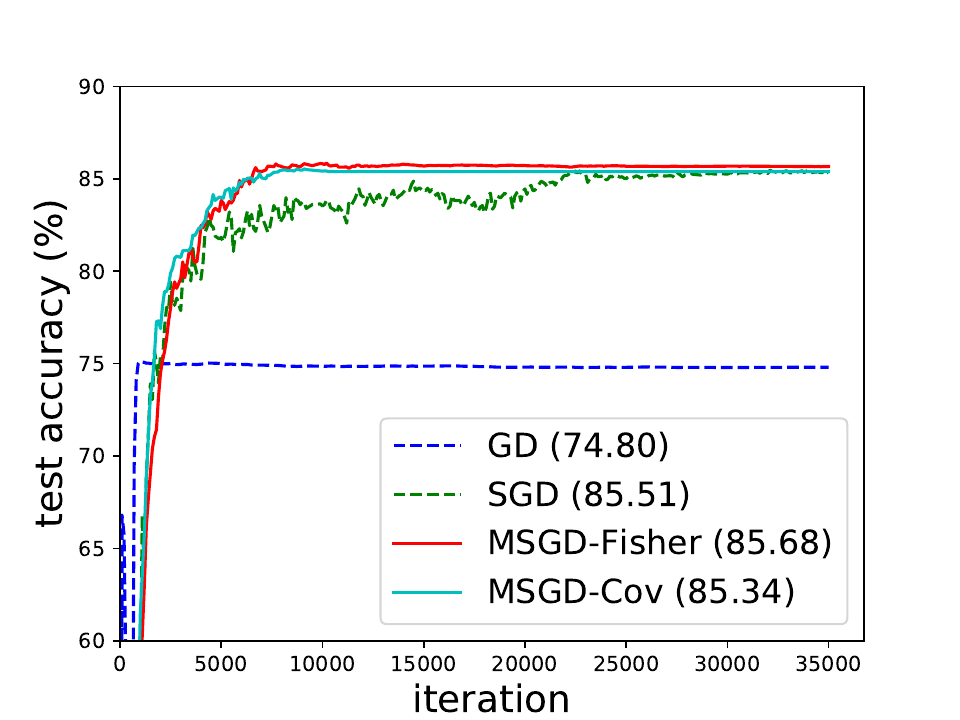} \\
(a) Small FashionMNIST & (b) Small SVHN & (c) CIFAR-10
\end{tabular}
\vspace{-0.15cm}
\caption{\small 
The generalization of MSGD.
X-axis: number of iterations; y-axis: test accuracy.
\textbf{(a)}: We randomly draw $1,000$ samples from FashionMNIST as the training set, then train a small convolutional network with them.
\textbf{(b)}: We use $25,000$ samples from SVHN as the training set, then train a VGG-11 without Batch Normalization.
\textbf{(c)}: We train a ResNet-18 on CIFAR-10 without using data augmentation and weight decay.
\textbf{MSGD-Fisher}: MSGD with Gaussian gradient noise whose covariance is the scaled Fisher.
\textbf{MSGD-Cov}: MSGD with Gaussian gradient noise whose covariance is the SGD covariance.
\textbf{MSGD-Bernoulli}: MSGD with Bernoulli sampling noise.
\textbf{MSGD-[Fisher-$\mathbf{B}$]}: MSGD-Fisher with the Fisher estimated using a mini-batch of samples in size $B$.
}
\label{fig:msgd}
\end{figure*}

\begin{rmk*}
In the traditional setting of machine learning, the number of samples is much larger than the number of parameters,  $d\ll n$.
And the SVD method for generating Gaussian noises is indeed plausible in this case.
However, when it comes to deep neural networks where $n\ll d$, it turns out computing the full gradient could be much cheaper than explicitly evaluating the covariance matrix and performing SVD. Thus for modern machine learning, our approach is far more efficient than the SVD method for injecting Gaussian noises with the SGD covariance.
\end{rmk*}

\textbf{Experiments}~
In Figure~\ref{fig:msgd} we test MSGD-Cov on various datasets and models.
The results consistently suggest that the MSGD-Cov can generalize well as the vanilla SGD, though its noise belongs to a different distribution class.
More interestingly, we observe that the MSGD-Cov converges faster than the vanilla SGD.

\subsection{Fisher vs. SGD covariance}\label{sec:fisher-cov}

In this part we discuss two kinds of commonly used Gaussian noises: the Gaussian noises with covariance as 
the SGD covariance, i.e., $\upsilon_{\text{C}}(\theta) \sim \Ncal (0, C(\theta))$~\cite{zhu2018anisotropic} 
and the scaled Fisher, i.e., $\upsilon_{\text{F}}(\theta) \sim \Ncal (0, \frac{1}{b}F(\theta))$, where $F(\theta) =  \frac{1}{n}\grad_{\theta}\Lcal(\theta)\grad_{\theta}\Lcal(\theta)^T$ is the Fisher.
We call the MSGD with these two noises the MSGD-Cov and the \emph{MSGD-Fisher}, respectively.
The two noises sometimes cause confusion in literature, since both of them are adopted for simulating the SGD noise~\cite{zhu2018anisotropic,wen2019interplay};
but we are not sure whether or not they have the same regularization effects~\cite{martens2014new,kunstner2019limitations,thomas2019interplay}.
The connection between the SGD covariance and the Fisher is clear: 
$C(\theta) = \frac{1}{b} (F(\theta) - \grad_{\theta}L(\theta) \grad_{\theta}L(\theta)^T)$, i.e., ignoring a factor of scaling, $C(\theta)$ is the second central moment of the SGD noise, while $F(\theta)$ is the second raw moment.
Next we discuss their common ground on imposing regularization.

Intuitively the two dynamics should not be far away from each other. 
We can see this by investigating the MSGD iteration~\eqref{eq:msgd}.
At the early phase of the training, the gradient term is much larger than the noise term in scale~\cite{shwartz2017opening} and dominates the optimization.
Thus the noise term almost makes no contribution, no matter whether its covariance is the SGD covariance or the scaled Fisher. 
During the latter phase, however, the gradient turns to be close to zero, thus $C(\theta) \approx \frac{1}{b}F(\theta)$ and $\upsilon_{\text{C}}(\theta) \approx \upsilon_{\text{F}}(\theta)$.
However, by such discussion neither the approximation is clear nor do we know about the transition phase.

Thanks to the sampling noise, we are able to develop a mathematical equivalence between the two noises along the whole training phase.
Let $\Vcal_{\text{C}}$ and $\Vcal_{\text{F}}$ be the sampling noises for $\upsilon_{\text{C}}(\theta)$ and $\upsilon_{\text{F}}(\theta)$ respectively, i.e., $\upsilon_{\text{C}}(\theta) = \grad_{\theta}\Lcal(\theta) \Vcal_{\text{C}}$ and $\upsilon_{\text{F}}(\theta) = \grad_{\theta}\Lcal(\theta) \Vcal_{\text{F}}$.
By the MSGD algorithm we have
\begin{align*}
    \Vcal_{\text{C}} 
    =& \frac{1}{\sqrt{bn}}\left(I-\frac{1}{n}\mathbbm{1}\mathbbm{1}^T\right)\epsilon
    = \frac{1}{\sqrt{bn}} \epsilon - \frac{1}{\sqrt{bn}} \frac{\mathbbm{1}^T\epsilon}{n}\mathbbm{1},\\
    \Vcal_{\text{F}} 
    =& \frac{1}{\sqrt{bn}} \epsilon,\qquad \epsilon \sim \Ncal(0,I_{n\times n}).
\end{align*}
Note the matrix $\left( I - \frac{1}{N}\mathbbm{1}\mathbbm{1}^T\right)$ centralizes a random vector.
But the components of the white noise $\epsilon$ are already i.i.d. of zero mean, thus $\frac{\mathbbm{1}^T\epsilon}{n} \approx 0$ by the law of large numbers.
Hence $\Vcal_{\text{C}} \approx \Vcal_{\text{F}}$ and $\upsilon_{\text{C}}(\theta) \approx \upsilon_{\text{F}}(\theta)$.
Moreover, the equivalence holds no matter where the parameter $\theta$ is, thanks to the fact that the sampling noises are state-independent. 
We conclude that the Fisher Gaussian noise and the SGD covariance noise must lead to identical regularization effect for learning deep models.

\textbf{Experiments}~
In Figure~\ref{fig:msgd} we present the experimental results regards MSGD-Cov and MSGD-Fisher.
Consistent with our analysis, the behavior of the MSGD-Fisher perfectly approximates that of the MSGD-Cov.
Hence the equivalence between the Fisher noise and the SGD covariance noise from the Gaussian class has been verified from both theory and experiments.
In the following study, we focus on MSGD-Fisher as the representative of the algorithms with noises from the Gaussian class.

\subsection{Bernoulli sampling noise}
Notice that the Fisher sampling noise $\Vcal_{\text{F}}$ has i.i.d. components and loses the covariance structure of the SGD sampling noise.
Nonetheless it can still regularize GD well (see MSGD-Fisher in Figure~\ref{fig:msgd}).
It suggests that a sampling noise with independent components is capable enough for imposing regularization.

To further verify this conjecture, we consider a \emph{Bernoulli sampling noise}:
$\Vcal_{\text{B}} = (v_1,\dots,v_n)^T$, where the components are i.i.d. and $\Pbb(v_i=\frac{1}{b}-\frac{1}{n}) = \frac{b}{n}, \Pbb(v_i=-\frac{1}{n}) = \frac{n-b}{n}$.
Then $\expect{\Vcal_{\text{B}}} = 0$ and $\Var [\Vcal_{\text{B}}] = \frac{n-b}{b n^2}I = \diag \bracket{\var{\Vcal_{\sgd}}}$, i.e., the covariance of the Bernoulli sampling noise is exactly the diagonal of the covariance of SGD sampling noise.
The Bernuolli sampling noise can also be easily injected to GD by Algorithm~\ref{alg:msgd}, and we call such algorithm the \emph{MSGD-Bernoulli}.

\textbf{Experiments}~
In Figure~\ref{fig:msgd} we find the MSGD-Bernoulli and the MSGD-Fisher both generalize as the vanilla SGD.
Thus sampling noises with independent components do not lose the regularization ability.
In contrast, gradient noises with independent components can never recover the regularization effects of the SGD noise.
For example one can look at the performance of \emph{GLD diag} in~\cite{zhu2018anisotropic}.
This comparison reveals a fundamental advantage of understanding the gradient noise from its sampling noise.

\begin{figure*}
\centering
\begin{tabular}{ccc}
\includegraphics[width=0.3\linewidth]{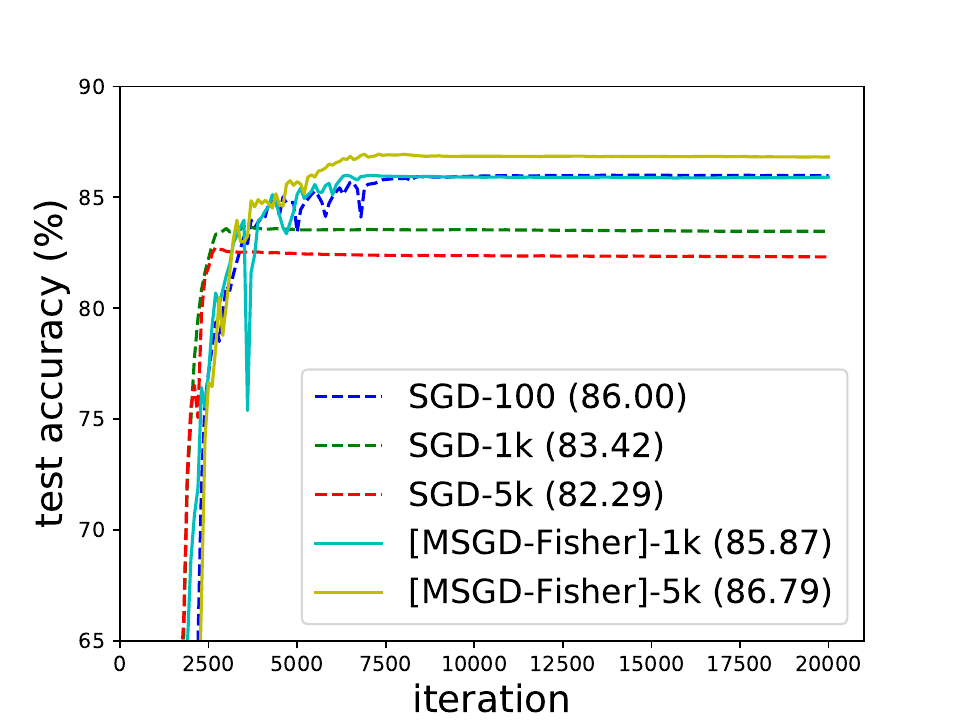} &
\includegraphics[width=0.3\linewidth]{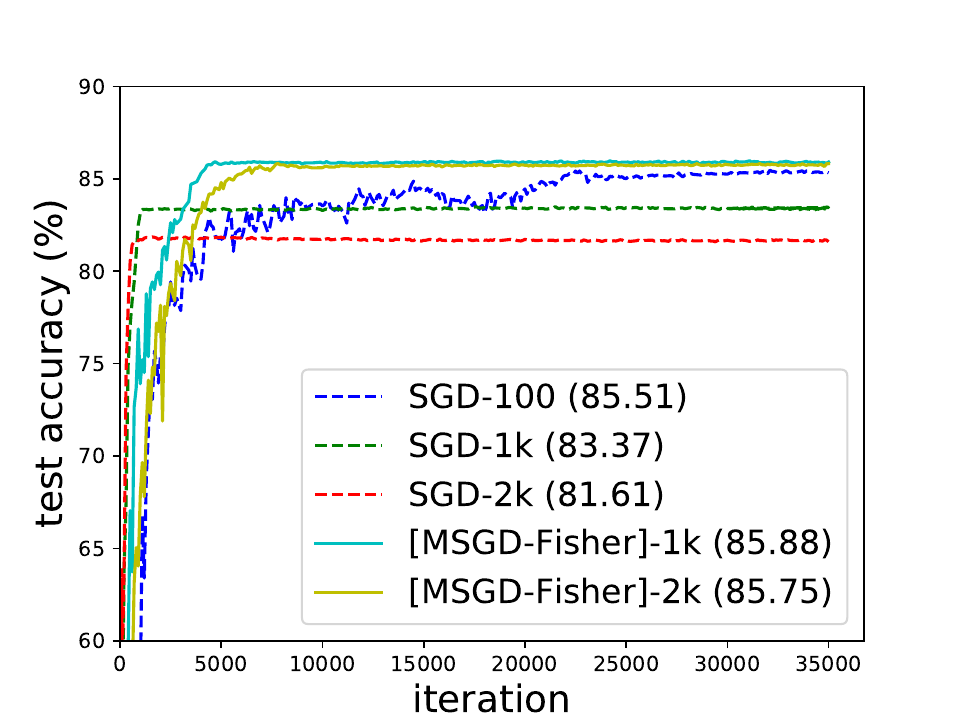} &
\includegraphics[width=0.3\linewidth]{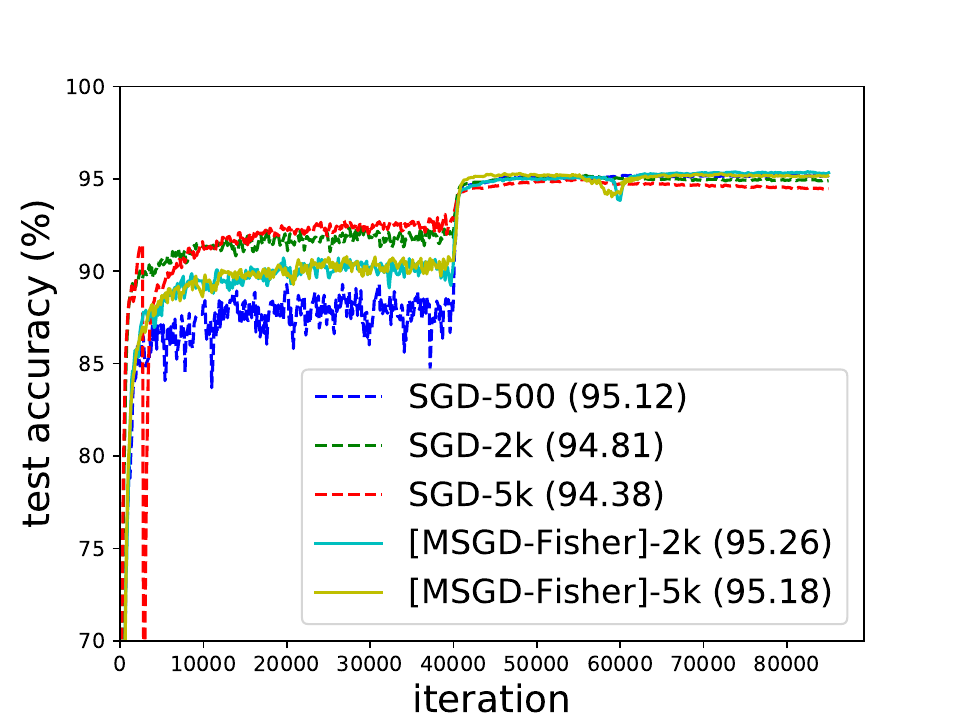} \\
(a) Small SVHN & (b) CIFAR-10 & (c) CIFAR-10 Standard 
\end{tabular}
\vspace{-0.15cm}
\caption{\small
The generalization of mini-batch MSGD.
X-axis: number of iterations; y-axis: test accuracy.
\textbf{(a)}: We use $25,000$ samples from SVHN as the training set, then train a VGG-11 without Batch Normalization.
\textbf{(b)}: We train a ResNet-18 on CIFAR-10 without using data augmentation and weight decay.
\textbf{(c)}: We train a ResNet-18 on CIFAR-10 with full tricks.
\textbf{SGD-$\mathbf{B}$}: SGD with batch size $B$.
\textbf{[MSGD-Fisher]-$\mathbf{B}$}: mini-batch MSGD with batch size $B$, and an compensatory sampling noise from the (sparse) Gaussian class.
}
\label{fig:batch-msgd}
\end{figure*}

\subsection{Sparse Gaussian sampling noises}

We then study another class of gradient noise who has \emph{sparse Gaussian sampling noise}.
The gradient noise is constructed as below:
we first draw a mini-batch of samples uniformly at random in size $B$, then estimate the Fisher using this mini-batch, then generate a Gaussian noise using the estimated Fisher as covariance, finally the noise is properly scaled to maintain the magnitude.
By Algorithm~\ref{alg:msgd}, the sampling noise is generated as
\begin{equation*}
    \Vcal_{F}' = \sqrt{B/b}\cdot \Vcal_{\sgd}(B)\odot \epsilon,
\end{equation*}
where $\Vcal_{\sgd}(B)$ is the SGD sampling noise with batch size $B$, and $\epsilon\in \Rbb^{n}$ is a white noise.
MSGD using $\Vcal_{F}'$ as sampling noise is denoted as \emph{MSGD-[Fisher-$B$]}, where $B$ is the batch size.
Note that $\Ebb[\Vcal_{F}'] = 0$ and $\Var[\Vcal_{F}'] = \Var[\Vcal_{\sgd}(b)]$, i.e., the sampling noise has the same magnitude and covariance structure as the SGD sampling noise.
Because $\Vcal_{F}'$ is a sparse Gaussian noise, its gradient noise belongs to neither the Gaussian class nor the SGD noise class.

\textbf{Experiments}~
The performance of MSGD-[Fisher-$B$] is shown in Figures~\ref{fig:msgd}.
Even with a very small batch size, e.g., $10$ for FashionMNIST and  $100$ for SVHN, MSGD-[Fisher-$B$] can generalize as MSGD-Fisher and SGD.
These results further support our understanding that the noise class is not the crux for regularization.

\subsection{Mini-batch MSGD}

Finally, we discuss the mini-batch version of MSGD which is of practical interests.
During each iteration of the vanilla MSGD, the information of full training set is required, which is unacceptable in practice.
As an extension, we introduce Algorithm~\ref{alg:batch-msgd}, the \emph{mini-batch MSGD}.
For example, when the plugged noise is Fisher Gaussian noise, we call the algorithm \emph{[MSGD-Fisher]-$B$}, where $B$ denotes the batch size of the mini-batch MSGD algorithm. 
We emphasize that the sampling noise in [MSGD-Fisher]-$B$ is a sparse Gaussian noise plus an SGD sampling noise, thus it belongs to a new class different from what we have discussed before.
However, thanks to the fact that noise class is unimportant the regularization ability, the noises we adopt here do not limit the capability of the mini-batch MSGD.

\begin{algorithm}
\caption{Mini-Batch Multiplicative SGD}
\label{alg:batch-msgd}
\begin{algorithmic}[1]
\STATE \textbf{Input}: Initial parameter $\theta_0\in \mathbb{R}^d$, training data $\{(x_i, y_i)\}_{i=1}^n$, loss function $\ell_i(\theta)=\ell((x_i, y_i), \theta)\in \Rbb$, learning rate $\eta>0$, batch size $b$
\FOR {$k=0,1,2,...,K-1$}
    \STATE Uniformly sample a mini-batch $ \{k_1,\dots, k_b\}$ and collect the loss vector $\Lcal(\theta_k) = \left(\ell_{k_1}(\theta_k),\dots,\ell_{k_b}(\theta_k)\right) \in \Rbb^{1\times b}$
    \STATE Generate a sampling noise $\Vcal\in \Rbb^b$ of zero mean and desired covariance
    \STATE Compute the sampling vector $\Wcal = \frac{1}{b}\mathbbm{1} + \Vcal$
    \STATE Calculate the randomized loss $\tilde{L}(\theta_k) = \Lcal(\theta_k) \Wcal$
    \STATE Compute the stochastic gradient $\grad_{\theta} \tilde{L}(\theta_k)$
    \STATE Update the parameter $\theta_{k+1} = \theta_k-\eta \grad_{\theta}\tilde{L}(\theta_k)$
\ENDFOR
\STATE \textbf{Output}: Output $\theta_K$
\end{algorithmic}
\end{algorithm}

\textbf{Large batch training}~
When training with SGD, as the batch size becomes large, the generalization gets hurt since the gradient noise tends to be small ~\cite{keskar2016large}. 
A promising method to close the generalization gap of large batch training is adding a compensatory gradient noise, e.g., a Gaussian gradient noise using scaled Fisher as covariance~\cite{wen2019interplay}.
However as we have discussed in Section~\ref{sec:gen-gauss-noise}, it is computationally costly to directly insert a structured Gaussian noise via SVD.
Instead, the algorithm [MSGD-Fisher]-$B$ provides an efficient method for injecting a compensatory sampling noise from the (sparse) Gaussian class.

\textbf{Experiments}~
We thus perform large batch training experiments with our [MSGD-Fisher]-$B$ algorithm.
The results are shown in Figure~\ref{fig:batch-msgd}.
Since in this case the covariance of the sampling noise becomes hard to calculate, we simply tune the noise magnitude to achieve its best performance.
As illustrated in Figure~\ref{fig:batch-msgd} (a) (b), on toy datasets the [MSGD-Fisher]-$B$ with large batch size has a even better generalization compared with small batch SGD.
Its convergence is also faster.
Even in real settings of training ResNet-18 on CIFAR10, Figure~\ref{fig:batch-msgd} (c) demonstrates that the [MSGD-Fisher]-$B$ with large batch size generalizes well as the small batch SGD, while SGD with large batch size performs worse.

\subsection{Empirical studies summary}
In Figure~\ref{fig:msgd} we compare the generalization performance of noisy gradient descents with noises from various different classes.
We find that, provide suitable magnitude and covariance structure, all the concerned noises can regularize gradient descent as the SGD noise.
These empirical results together with the theoretical evidence verify our understanding that the noise class is not a crux for regularization.
An interesting additional finding is that Gaussian MSGD tends to converge faster than others. 

In Figure~\ref{fig:batch-msgd} we present the empirical results of the mini-batch MSGD (Algorithm~\ref{alg:batch-msgd}).
Our algorithm perfectly closes the generalization gap of large batch training by injecting compensatory (sparse) Gaussian sampling noises.
Besides, our algorithm achieves this effect in a more efficient manner than the traditional way of inserting Gaussian gradient noise based on SVD.
These results demonstrate the promising application of the mini-batch MSGD algorithm in practice.

\section{Discussion}
\textbf{Benefits of Gaussian gradient noise}~
The continuous stochastic differential equations (SDEs) have been widely used for approximating and analyzing the discrete SGD iterations~\cite{li2017stochastic,hu2017diffusion,orvieto2019continuous}.
For SGD, this continuous approximation only hold in weak sense~\cite{li2017stochastic}.
For Gaussian MSGD, however, a strong convergence can be established between the discrete iterations and the continuous SDEs.
This is discussed more in Supplementary Materials, Section~\ref{sec:pf-strong-converge}.
The strong convergence guarantees a path-wise closeness between the discrete iterations and the continuous paths, 
beyond the close behavior at the level of probability distributions guaranteed by weak convergence.
This advantage of Gaussian MSGD might account for its observed faster convergence.

\textbf{The importance of the gradient matrix $\grad_{\theta} \Lcal (\theta)$}~
Consider the MSGD-Bernoulli/Fisher and the \emph{GLD diag} from~\cite{zhu2018anisotropic},
empirical studies show that the MSGD-Bernoulli/Fisher generalize well as SGD, while the GLD diag performs much worse.
In the MSGD-Bernoulli/Fisher the sampling noises have independent components, and in the GLD diag the gradient noise has independent components.
Though the compared algorithms all discard certain ``dependece'' in their noises, the MSGD-Bernoulli/Fisher keep the full information of the gradient matrix, while the GLD diag severally destroys its structure.
We thus conjecture that the gradient matrix contains key information for the regularization induced by noises.

\section{Conclusion}
In this work we introduce a novel kind of gradient noise as the composition of the gradient matrix and a sampling noise, which includes the SGD noise.
By investigating these noises we find the noise class is not a crux for regularization, provided suitable noise magnitude and covariance structure.
Furthermore, we show that the scaled Fisher and the gradient covariance of SGD is equivalent when serve as the covariance of noises from the Gaussian class.
Finally, an algorithm is proposed to perform noisy gradient descent that generalizes as SGD. The algorithm can be extended for practical usage like large batch training.

\section*{Acknowledgement}

This project is supported by National Natural Science Foundation of China  (No.61806009 and 61932001), PKU-Baidu Funding 2019BD005, Beijing Academy of Artificial Intelligence (BAAI),  Intelligent Manufacturing Action Plan of Industrial Solid Foundation Program (JCKY2018204C004), NSF CAREER grant 1652257, ONR Award
N00014-18-1-2364, and the Lifelong Learning Machines program from DARPA/MTO.

\bibliography{references}
\bibliographystyle{icml2020}

\onecolumn
\appendix

\section{Missing proofs in main paper}
\subsection{Proof of Proposition~\ref{thm:sampling-noise}}\label{sec:pf-sampling-noise}

\begin{proof}
We first calculate the expectation and variance of the sampling random vector $\Wcal_{\sgd}$, then obtain that of the sampling noise $\Vcal_{\sgd}$.

\paragraph{Sampling with replacement}
In the circumstance of sampling with replacement, the sampling random vector $\Wcal_{\sgd}$ could be decompose as
\begin{equation*}
    \Wcal_{\sgd} = \Wcal^1 + \dots + \Wcal^b,
\end{equation*}
where $\Wcal^1,\dots,\Wcal^b$ are i.i.d. and each of them represents once sampling procedure.
Thus $\Wcal^i = (w^i_1, \dots, w^i_n)^T$ contains one multiple of $\frac{1}{b}$ and $n-1$ multiples of zero, with random index.
Hence we have 
\begin{equation*}
    \Ebb [w^i_j] = \frac{1}{bn},\quad
    \Ebb [w^i_j w^i_j] = \frac{1}{b^2n},\quad
    \Ebb [w^i_j w^i_k] = 0,\ \forall j \ne k.
\end{equation*}
Thus
\begin{align*}
    \Ebb[\Wcal^i] =& \frac{1}{bn}\mathbbm{1}, \\
    \Var[\Wcal^i] =& \Ebb[\Wcal^i (\Wcal^i)^T] - \Ebb[\Wcal^i]\Ebb[\Wcal^i]^T
    =\begin{pmatrix}
    \frac{1}{b^2n} & & \\
     & \ddots & \\
     & & \frac{1}{b^2n}
    \end{pmatrix}
    - \frac{1}{b^2 n^2}\mathbbm{1}\mathbbm{1}^T
    = \frac{1}{b^2 n}\left(I - \frac{1}{n}\mathbbm{1}\mathbbm{1}^T\right).
\end{align*}
Recall $\Wcal^1,\dots,\Wcal^b$ are i.i.d., thus
\begin{equation*}
    \Ebb[\Wcal_{\sgd}] = b\Ebb[\Wcal^i] = \frac{1}{n}\mathbbm{1}, \quad
    \Var[\Wcal_{\sgd}] = b\Var[\Wcal^i] = \frac{1}{bn}\left(I - \frac{1}{n}\mathbbm{1}\mathbbm{1}^T\right).
\end{equation*}
Therefore for the sampling noise $\Vcal_{\sgd} = \Wcal_{\sgd} - \frac{1}{n}\mathbbm{1}$ we have
\begin{equation*}
    \Ebb[\Vcal_{\sgd}] = 0, \quad
    \Var[\Vcal_{\sgd}] = \frac{1}{bn}\left(I - \frac{1}{n}\mathbbm{1}\mathbbm{1}^T\right).
\end{equation*}

\paragraph{Sampling without replacement}
Let $\Wcal_{\sgd}^{\prime} = (w'_1,\dots,w'_n)^T$.
In the case of sampling without replacement, we know the sampling random vector $\Wcal_{\sgd}^{\prime}$ contains exactly $b$ multiples of $\frac{1}{b}$s and $n-b$ multiples of zero, with random index.
Hence we have
\begin{equation*}
    \Ebb[w'_j] = \frac{\binom{n-1}{b-1} \frac{1}{b}}{\binom{n}{b}} = \frac{1}{n}, \quad
    \Ebb[(w'_j)^2] = \frac{\binom{n-1}{b-1} \frac{1}{b^2}}{\binom{n}{b}} = \frac{1}{bn}, \quad
    \Ebb[w'_j w'_k] = \frac{\binom{n-2}{b-2} \frac{1}{b^2}}{\binom{n}{b}} = \frac{b-1}{bn(n-1)}, \ \forall j \ne k.
\end{equation*}
Thus
\begin{align*}
    \Ebb[\Wcal_{\sgd}^{\prime}] = & \frac{1}{N}\mathbbm{1}, \\
    \Var[\Wcal_{\sgd}^{\prime}] = & \Ebb[\Wcal_{\sgd}^{\prime} (\Wcal_{\sgd}^{\prime})^T] - \Ebb[\Wcal^i]\Ebb[\Wcal^i]^T \\
    =& \begin{pmatrix}
    \frac{1}{bn} & \frac{b-1}{bn(n-1)} & \cdots & \frac{b-1}{bn(n-1)} \\
    \frac{b-1}{bn(n-1)} & \frac{1}{bn} & \cdots & \frac{b-1}{bn(n-1)} \\
    \vdots & \vdots & \ddots & \vdots \\
    \frac{b-1}{bn(n-1)} & \frac{b-1}{bn(n-1)} & \cdots & \frac{1}{b^2 n}
    \end{pmatrix}
    - \frac{1}{n^2}\mathbbm{1}\mathbbm{1}^T
    = \frac{n-b}{bn(n-1)}\left(I - \frac{1}{n}\mathbbm{1}\mathbbm{1}^T\right).
\end{align*}
Therefore for the sampling noise $\Vcal_{\sgd}^{\prime} = \Wcal_{\sgd}^{\prime} - \frac{1}{n}\mathbbm{1}$ we have
\begin{equation*}
    \Ebb[\Vcal_{\sgd}^{\prime}] = 0, \quad
    \Var[\Vcal_{\sgd}^{\prime}] = \frac{n-b}{bn(n-1)}\left(I - \frac{1}{n}\mathbbm{1}\mathbbm{1}^T\right).
\end{equation*}

\end{proof}

\subsection{Proof of Theorem~\ref{thm:sgd-generalization}}\label{sec:pf-sgd-generalization}
\begin{proof}

Let 
\begin{equation*}
    \epsilon_n = y_n - x_n^T \theta_*,
\end{equation*}
by assumption we have
\begin{equation*}
    \Ebb [\epsilon_n x_n ] = 0,\quad 
    \Ebb[\epsilon_n] = 0,\quad \Ebb[\epsilon^2 x x^T]\preceq \sigma^2 \Sigma.
\end{equation*}
Recall the MSGD updates
\begin{equation*}
    \theta_{n+1} = \theta_n - \eta \sum_{r\in B_n} w_r \left( x_r x_r^T \theta_n - y_r x_r \right),
\end{equation*}
hence we have
\begin{equation*}
    \theta_{n+1} - \theta_* = \left( I - \eta \sum_{r\in B_n} w_r x_r x_r^T \right) (\theta_n - \theta_*) + \eta \sum_{r \in B_n} w_r \epsilon_r x_r.
\end{equation*}

Define 
\begin{align}
    & L(k) = \sum_{r\in B_k} w_r x_r x_r^T,\label{eq:defi-L} \\
    & M(i,k) =\begin{cases}
    &\left( I - \eta L(i) \right)  \cdots \left( I - \eta L(k) \right),\quad  i \ge k \\
    & I,\quad  i < k. \label{eq:defi-M}     
    \end{cases}\\
    & N(k) = \sum_{r\in B_k} w_r \epsilon_r x_r,\label{eq:defi-N}
\end{align}
Then recursively we obtain
\begin{equation}\label{eq:theta-solution}
    \theta_i - \theta_* = M(i,1) (\theta_0 - \theta_*) + \eta \sum_{k=1}^i M(i, k+1) N(k).
\end{equation}

\paragraph{Moments of $L(k)$}
We first calculate the first and second moments of $N(k)$ defined in Eq.~\eqref{eq:defi-N}.
Since $\Ebb[w_r] = \frac{1}{B},\ \Ebb[w_i^2] = \frac{1}{bB},\ \Ebb[w_i w_j] = \frac{b-1}{bB(B-1)},\ i \ne j$, and $\Ebb \left[\norm{x}_2^2 x x^T\right] \preceq R^2 \Sigma,\ \Sigma \preceq \lambda I$, we have
\begin{align*}
    \Ebb \left[L(k)\right] =& \sum_{r=1}^B \Ebb[w_r]\cdot \Ebb [x_r x_r^T] = B \cdot \frac{1}{B} \cdot \Sigma = \Sigma.\\
    \Ebb \left[L(k)^2\right]
    =& \Ebb \sum_{r=1}^B w_r^2 x_r x_r^T x_r x_r^T + 2\Ebb \sum_{r=1}^{B-1}\sum_{s=2}^{B} w_r w_s x_r x_r^T x_s x_s^T \\
    =& \sum_{r=1}^B \Ebb [w_r^2]\cdot \Ebb \left[\norm{x_r}_2^2 x_r x_r^T\right] + 2\sum_{r=1}^{B-1}\sum_{s=2}^{B} \Ebb [w_r w_s] \cdot \Ebb [x_r x_r^T]\cdot \Ebb [x_s x_s^T] \\
    =& B \cdot \frac{1}{bB} \cdot \Ebb \left[\norm{x}_2^2 x x^T\right] + 2 \frac{B(B-1)}{2}\cdot \frac{b-1}{bB(B-1)} \cdot \Sigma^2 \\
    \preceq & \frac{1}{b} R^2 \Sigma + \frac{b-1}{b} \lambda \Sigma 
    = \frac{R^2 + (b-1)\lambda}{b} \Sigma.
\end{align*}

\paragraph{Moments of $M(i,k)$}
We only consider $i\ge k$.
\begin{align*}
    \Ebb[M(i,k)] =& \left(I-\eta \Ebb[L(i)]\right)\cdots \left(I-\eta \Ebb[L(k)]\right)
    = (I-\eta\Sigma)^{i-k+1}.\\
    \Ebb M(i,k) M(i,k)^T 
    = & \Ebb M(i,k+1) \left( I - \eta L(k)\right)^2 M(i,k+1)^T \\
    = & \Ebb M(i,k+1) \left( I - 2\eta L(k) + \eta^2 L(k)^2 \right) M(i,k+1)^T \\
    \le & \Ebb M(i,k+1) \left( I - 2\eta \Sigma + \eta^2 \frac{R^2 + (b-1)\lambda}{b} \Sigma \right) M(i,k+1)^T \\
    = & \Ebb M(i,k+1)M(i,k+1)^T - \eta\left(2 - \eta \frac{R^2 + (b-1)\lambda}{b}\right) \Ebb M(i,k+1) \Sigma M(i,k+1)^T.
\end{align*}
Hence
\begin{equation*}
    \Ebb M(i,k+1) \Sigma M(i,k+1)^T \le \frac{1}{\eta\left(2 - \eta \frac{R^2 + (b-1)\lambda}{b}\right)}\left( \Ebb M(i,k+1)M(i,k+1)^T - \Ebb M(i,k) M(i,k)^T \right).
\end{equation*}

\paragraph{Moments of $N(k)$}
\begin{align*}
    \Ebb[N(k)] =& \sum_{r\in B_k} \Ebb[w_r] \cdot \Ebb[\epsilon_r x_r] = B \cdot \frac{1}{B} \cdot 0 = 0.\\
    \Ebb\left[ N(k) N(k)^T \right] 
    =& \Ebb \sum_{r=1}^B w_r^2 \epsilon_r^2 x_r x_r^T + 2\Ebb \sum_{r=1}^{B-1}\sum_{s=2}^B w_r w_s \epsilon_r \epsilon_s x_r x_s^T \\
    =& \sum_{r=1}^B \Ebb[w_r^2]\cdot \Ebb[\epsilon_r^2 x_r x_r^T] + 2\sum_{r=1}^{B-1}\sum_{s=2}^B \Ebb[w_r]\cdot \Ebb[w_s] \cdot\Ebb[\epsilon_r x_r]\cdot \Ebb[\epsilon_s x_s]^T \\
    =& B\cdot \frac{1}{bB}\cdot \Ebb[\epsilon^2 x x^T] + 2\frac{B(B-1)}{2}\cdot \frac{1}{B^2} \cdot 0\\
    \le & \frac{1}{b} \sigma^2 \Sigma.
\end{align*}

\paragraph{Calculate averaging}
Takeing expectation to $w_k$ and $B_k$, we have
\begin{align*}
    &\Ebb \sum_{i=0}^{n-1} \sum_{j=i+1}^n \left< \theta_i - \theta_*, \Sigma (\theta_j - \theta_*) \right> \\
    =& \Ebb \sum_{i=0}^{n-1} \sum_{j=i+1}^n \left< \theta_i - \theta_*, \Sigma \left( M(j,i+1)(\theta_i - \theta_*) + \eta \sum_{k=i+1}^j M(j,k+1) N(k) \right) \right> \\
    =& \Ebb \sum_{i=0}^{n-1} \sum_{j=i+1}^n \left< \theta_i - \theta_*, \Sigma  M(j,i+1)(\theta_i - \theta_*) \right> \\
    =& \Ebb \sum_{i=0}^{n-1} \sum_{j=i+1}^n \left< \theta_i - \theta_*, \Sigma \left( I-\eta\Sigma \right)^{j-i} (\theta_i - \theta_*) \right> \\
    =& \Ebb \sum_{i=0}^{n-1} \left< \theta_i - \theta_*, \eta^{-1} \left( I-\eta\Sigma - (I-\eta \Sigma)^{n-i+1} \right)(\theta_i - \theta_*) \right> \\
    \le & \Ebb \sum_{i=0}^{n} \left< \theta_i - \theta_*, \eta^{-1} \left( I-\eta\Sigma \right)(\theta_i - \theta_*) \right> \\
    = & \eta^{-1} \Ebb \sum_{i=0}^n \norm{\theta_i - \theta_*}_2^2 - \Ebb \sum_{i=0}^n \norm{\Sigma^{\half} (\theta_i - \theta_*)}_2^2,
\end{align*}
which implies that
\begin{align*}
    & (n+1)^2 \Ebb\norm{\Sigma^{\half} (\bar{\theta}_n - \theta_*)}_2^2 = \Ebb \sum_{i,j=0}^n \left< \theta_i-\theta_*, \Sigma(\theta_j-\theta_*) \right> \\
    = & \Ebb \sum_{i=0}^n \left< \theta_i-\theta_*, \Sigma (\theta_i-\theta_*) \right> + 2 \Ebb \sum_{i=0}^{n-1} \sum_{j=i+1}^n \left< \theta_i-\theta_*, \Sigma (\theta_j-\theta_*) \right> \\
    \le & \Ebb \sum_{i=0}^n \norm{\Sigma^{\half} (\theta_i - \theta_*)}_2^2 + 2 \eta^{-1} \Ebb \sum_{i=0}^n \norm{\theta_i - \theta_*}_2^2 - 2 \Ebb \sum_{i=0}^n \norm{\Sigma^{\half} (\theta_i - \theta_*)}_2^2 \\
    \le & 2 \eta^{-1} \Ebb \sum_{i=0}^n \norm{\theta_i - \theta_*}_2^2,
\end{align*}
and in the following we bound $\Ebb \sum_{i=0}^n \norm{\theta_i - \theta_*}_2^2$.
We do so by bounding each term.

Now since the solution of $\theta_i$ in Eq.~\eqref{eq:theta-solution} and the fact $\Ebb[N(k)]=0$, we have
\begin{align*}
    \Ebb\norm{\theta_i - \theta_*}_2^2 
    =& \Ebb\norm{M(i,1)(\theta_0-\theta_*)}_2^2 + \eta^2 \Ebb \sum_{k=1}^i\sum_{j=1}^i \left< M(i,k+1) N(k),\ M(i,j+1) N(j)\right> \\
    =& \Ebb\norm{M(i,1)(\theta_0-\theta_*)}_2^2 + \eta^2 \Ebb \sum_{k=1}^i \left< M(i,k+1) N(k),\ M(i,k+1) N(k)\right>.
\end{align*}

In conclusion we have
\begin{align*}
    & \half \eta (n+1)^2 \Ebb\norm{\Sigma^{\half} (\bar{\theta}_n - \theta_*)}_2^2
    \le \Ebb \sum_{i=0}^n \norm{\theta_i - \theta_*}_2^2 \\
    = & \Ebb \sum_{i=0}^n \norm{M(i,1)(\theta_0-\theta_*)}_2^2 + \eta^2 \Ebb \sum_{i=0}^n \sum_{k=1}^i \left< M(i,k+1) N(k),\ M(i,k+1) N(k)\right>.
\end{align*}
We call the two terms as the noiseless term and the noise term.

\paragraph{Noise term}
We bound the noise term by observing that
\begin{align*}
    &\Ebb \left< M(i,k+1) N(k),\ M(i,k+1) N(k)\right> \\
    =& \Ebb M(i,k+1) N(k) N(k)^T M(i,k+1)^T \\
    =& \Ebb \tr \left[ N(k) N(k)^T M(i,k+1)^T M(i,k+1)\right] \\
    =& \tr \left[\Ebb \left[  N(k) N(k)^T \right] \cdot \Ebb \left[ M(i,k+1)^T M(i,k+1) \right] \right] \\
    \preceq & \sigma^2 \tr \left[ \Sigma \cdot \Ebb \left[ M(i,k+1)^T M(i,k+1) \right] \right] \\
    =& \sigma^2 \tr \Ebb \left[ M(i,k+1)\Sigma M(i,k+1)^T \right] \\
    \le & \frac{\sigma^2}{\eta\left(2 - \eta \frac{R^2 + (b-1)\lambda}{b}\right)} \tr \left( \Ebb M(i,k+1)M(i,k+1)^T - \Ebb M(i,k) M(i,k)^T \right).
\end{align*}
Hence
\begin{align*}
    & \eta^2 \Ebb \sum_{i=0}^n \sum_{k=1}^i \left< M(i,k+1) N(k),\ M(i,k+1) N(k)\right> \\
    \le & \frac{\eta \sigma^2}{\left(2 - \eta \frac{R^2 + (b-1)\lambda}{b}\right)} \sum_{i=0}^n \sum_{k=1}^i \tr \left( \Ebb M(i,k+1)M(i,k+1)^T - \Ebb M(i,k) M(i,k)^T \right) \\
    = & \frac{\eta \sigma^2}{\left(2 - \eta \frac{R^2 + (b-1)\lambda}{b}\right)} \sum_{i=0}^n \tr \left( \Ebb M(i,i+1)M(i,i+1)^T - \Ebb M(i,1) M(i,1)^T \right) \\
    \le & \frac{\eta \sigma^2}{\left(2 - \eta \frac{R^2 + (b-1)\lambda}{b}\right)} \sum_{i=0}^n \tr \left( \Ebb M(i,i+1)M(i,i+1)^T \right) \\
    = & \frac{\eta \sigma^2}{\left(2 - \eta \frac{R^2 + (b-1)\lambda}{b}\right)}(n+1) \tr [I] 
    = \frac{\eta \sigma^2}{\left(2 - \eta \frac{R^2 + (b-1)\lambda}{b}\right)}(n+1)d.
\end{align*}

\paragraph{Noiseless term}

Let $E_0 = (\theta_0-\theta_*)(\theta_0-\theta_*)^T$.
Define two linear operators $S$ and $T$ from symmetric matrices to symmetric matrices as
\begin{align*}
    S A =& \Ebb \left[ L(k) A L(k) \right] \\
    T A =& \Sigma A + A \Sigma - \eta \Ebb \left[ L(k) A L(k) \right] = \Sigma A + A \Sigma - \eta S A.
\end{align*}
With these notations and $M(i,1) = (I-\eta L(i))\cdots (I-\eta L(1))$, we recursively have
\begin{equation*}
    \Ebb \left[ M(i,1)^T M(i,1) \right] = (I-\eta T)^i I.
\end{equation*}

Next we bound the noiseless term
\begin{align*}
    \Ebb \sum_{i=0}^n \norm{M(i,1)(\theta_0-\theta_*)}_2^2 
    =& \Ebb \sum_{i=0}^n \tr \left[ M(i,1)^T M(i,1) (\theta_0-\theta_*)(\theta_0-\theta_*)^T \right] \\
    =& \sum_{i=0}^n \left< \Ebb M(i,1)^T M(i,1), E_0 \right> \\
    =& \sum_{i=0}^n \left< (I-\eta T)^i I, E_0 \right> \\
    =& \left< \eta^{-1}T^{-1} \left( I - (I-\eta T)^{n+1} \right) I, E_0 \right> \\
    \le & \left< \eta^{-1}T^{-1} I, E_0 \right>.
\end{align*}
Let $M = T^{-1} I$, then $I = TM = \Sigma M + M \Sigma - \eta S M$, hence by the Kronecker's produce we have
\begin{equation*}
    I + \eta S M = \Sigma M + M \Sigma = \left( \Sigma \otimes I + I \otimes \Sigma \right) M,
\end{equation*}
thus
\begin{equation*}
    M = \left( \Sigma \otimes I + I \otimes \Sigma \right)^{-1} I + \left( \Sigma \otimes I + I \otimes \Sigma \right)^{-1}\eta S M = \half \Sigma^{-1} + \left( \Sigma \otimes I + I \otimes \Sigma \right)^{-1}\eta S M.
\end{equation*}
Therefore
\begin{align*}
    & \Ebb \sum_{i=0}^n \norm{M(i,1)(\theta_0-\theta_*)}_2^2
    = \left< \eta^{-1}M, E_0 \right> = \frac{1}{2\eta} \left< \Sigma^{-1}, E_0 \right> + \left< \left( \Sigma \otimes I + I \otimes \Sigma \right)^{-1} S M , E_0 \right> \\
    =& \frac{1}{2\eta} (\theta_0 - \theta_*)^T \Sigma^{-1} (\theta_0 - \theta_*) + \left< S M , \left( \Sigma \otimes I + I \otimes \Sigma \right)^{-1} E_0 \right>.
\end{align*}
We left to bound $SM$ and $\left( \Sigma \otimes I + I \otimes \Sigma \right)^{-1} E_0$.

\subparagraph{Bound $\left( \Sigma \otimes I + I \otimes \Sigma \right)^{-1} E_0$}
By Cauchy-Schwarz inequality we have
\begin{equation*}
    E_0 = \Sigma^{\half} \Sigma^{-\half} (\theta_0 - \theta_*) (\theta_0 - \theta_*)^T \Sigma^{-\half} \Sigma^{\half} \preceq (\theta_0 - \theta_*)^T \Sigma^{-1} (\theta_0 - \theta_*) \cdot \Sigma.
\end{equation*}
Thus
\begin{equation*}
    \left( \Sigma \otimes I + I \otimes \Sigma \right)^{-1} E_0 
    \preceq (\theta_0 - \theta_*)^T \Sigma^{-1} (\theta_0 - \theta_*) \cdot  \left( \Sigma \otimes I + I \otimes \Sigma \right)^{-1}\Sigma 
    = (\theta_0 - \theta_*)^T \Sigma^{-1} (\theta_0 - \theta_*) \cdot \half I.
\end{equation*}
\subparagraph{Bound $SM$}
Firstly by definition,
\begin{align*}
    \tr [SM] =& \Ebb \tr \left[L(k)ML(k)\right]
    = \Ebb \sum_{r=1}^B \tr [w_r^2 x_r x_r^T M x_r x_r^T] + 2\Ebb \sum_{r=1}^{B-1}\sum_{s=2}^{B} \tr [w_r w_s x_r x_r^T M x_s x_s^T] \\
    =& \sum_{r=1}^B \Ebb [w_r^2]\cdot  \tr \left[\Ebb \left[\norm{x_r}_2^2 x_r x_r^T\right]M\right] + 2\sum_{r=1}^{B-1}\sum_{s=2}^{B}  \Ebb [w_r w_s] \cdot\tr\left[ \Ebb [x_r x_r^T]\cdot M \cdot \Ebb [x_s x_s^T] \right]\\
    \le & B \cdot \frac{1}{bB} \cdot \tr\left[R^2 \Sigma M\right] + 2 \frac{B(B-1)}{2}\cdot \frac{b-1}{bB(B-1)} \cdot \tr\left[\Sigma M \Sigma\right] \\
    \le & \frac{R^2 }{b} \cdot \tr \left[\Sigma M\right] + \frac{b-1}{b} \cdot \lambda \cdot \tr \left[ M \Sigma\right]
    = \frac{R^2 + (b-1)\lambda}{b} \tr \left[ \Sigma M\right].
\end{align*}
Secondly taking trace we have
\begin{equation*}
    d = \tr [I] = \tr [TM] = 2\tr [\Sigma M] - \eta \tr [SM] \ge 2\tr [\Sigma M] \ge  \frac{2b}{R^2 + (b-1)\lambda} \tr [SM],
\end{equation*}
which implies that $\tr [SM] \le \frac{R^2 + (b-1)\lambda}{2b}d$.

To sum up we have
\begin{align*}
    & \left< S M ,\ \left( \Sigma \otimes I + I \otimes \Sigma \right)^{-1} E_0 \right> \\
    \le & \half (\theta_0 - \theta_*)^T \Sigma^{-1} (\theta_0 - \theta_*) \cdot \left< S M , I \right> \\
    =& \half (\theta_0 - \theta_*)^T \Sigma^{-1} (\theta_0 - \theta_*) \cdot \tr [S M]\\
    \le & \frac{(R^2 + (b-1)\lambda) d}{4b}(\theta_0 - \theta_*)^T \Sigma^{-1} (\theta_0 - \theta_*).
\end{align*}
Therefore for the noiseless term we have
\begin{align*}
    & \Ebb \sum_{i=0}^n \norm{M(i,1)(\theta_0-\theta_*)}_2^2
    = \frac{1}{2\eta} (\theta_0 - \theta_*)^T \Sigma^{-1} (\theta_0 - \theta_*) + \left< S M , \left( \Sigma \otimes I + I \otimes \Sigma \right)^{-1} E_0 \right> \\
    \le & \left( \frac{1}{2\eta} + \frac{(R^2 + (b-1)\lambda) d}{4b}\right) (\theta_0 - \theta_*)^T \Sigma^{-1} (\theta_0 - \theta_*).
\end{align*}

In conclusion we have
\begin{align*}
    & \half \eta (n+1)^2 \Ebb\norm{\Sigma^{\half} (\bar{\theta}_n - \theta_*)}_2^2 
    \le \text{noiseless term} + \text{noise term} \\
    \le & \frac{\eta \sigma^2}{\left(2 - \eta \frac{R^2 + (b-1)\lambda}{b}\right)}(n+1)d + \left( \frac{1}{2\eta} + \frac{(R^2 + (b-1)\lambda) d}{4b}\right) (\theta_0 - \theta_*)^T \Sigma^{-1} (\theta_0 - \theta_*).
\end{align*}
Hence
\begin{equation*}
    \Ebb\norm{\Sigma^{\half} (\bar{\theta}_n - \theta_*)}_2^2 
    \le \frac{1}{n+1}\cdot \frac{2\sigma^2 d}{\left(2 - \eta \frac{R^2 + (b-1)\lambda}{b}\right)} 
    + \frac{1}{(n+1)^2}\cdot \left( 1 + \frac{(R^2 + (b-1)\lambda) \eta d}{2b}\right) (\theta_0 - \theta_*)^T \Sigma^{-1} (\theta_0 - \theta_*),
\end{equation*}
which complete our proof.

\end{proof}

\section{Strong convergence of Gaussian MSGD and its SDE}\label{sec:pf-strong-converge}
\begin{thm}(Strong convergence between Gaussian MSGD and SDE)\label{thm:strong-convergence}
Let $T \ge 0$. Let $C(\theta)$ be the diffusion matrix, e.g.,
$C(\theta) = \frac{1}{\sqrt{bN}}\grad_{\theta}\Lcal(\theta) \in \Rbb^{D\times N}$. Assume there exist some $L, M>0$ such that  $\max\limits_{i=1,2,...,N}(|\nabla_\theta \ell_i(\theta)|)\leq M$ and that $\nabla \ell_i(\theta)$ are Lipschitz continuous with bounded Lipschitz constant $L>0$ uniformly for all $i=1,2,...,N$.

Then the Gaussian MSGD iteration~\eqref{eq:M-SGD-gaussian}
\begin{equation}\label{eq:M-SGD-gaussian}
\theta_{k+1}-\theta_k = -\eta \nabla_{\theta} L(\theta_k) + \eta C(\theta_k)\Wcal_{k+1}, \ \Wcal_{k} \sim \Ncal(0,I),\ i.i.d.
\end{equation}
is a order $1$ strong approximation to SDE~\eqref{eq:sde}
\begin{equation}\label{eq:sde}
\dif \Theta_t = - \nabla_{\theta}L(\Theta_t) \dif t+\sqrt{\eta} C(\Theta_t) \dif W_t,\ \Theta_0=\theta_0, \ \text{$W_t\in \Rbb^N$ is a standard Brownian motion}
\end{equation}
i.e., there exist a constant $C$ independent on $\eta$ but depending on $L$ and $M$ such that
\begin{equation}
    \Ebb \| \Theta_{k\eta} - \theta_k \|^2 \le C \eta^2, \quad \text{ for all } 0\leq k\leq \lfloor T/\eta \rfloor.
\end{equation}
\end{thm}
\begin{proof}
We show that, as $\eta \rightarrow 0$, the discrete iteration $\theta_k$ of Eq.~\eqref{eq:M-SGD-gaussian} in strong norm and on finite--time intervals is close to the solution of the SDE~\eqref{eq:sde}.
The main techniques follow~\cite{borkar1999strong}, but~\cite{borkar1999strong} only considered the case when $C(\theta)$ is a constant.

For vector $x\in\Rbb^d$, we define its norm as $|x|:=\sqrt{x^T x}$; for matrix $X\in\Rbb^{d_1\times d_2}$, we define its norm as $|X|:=\sqrt{\tr (X^T X)}=\sqrt{\tr (XX^T)}$.

Let $\widehat{\Theta}_t$ be the process defined by the integral form of the stochastic differential equation

\begin{equation}\label{Eq:HatTheta}
\widehat{\Theta}_t-\widehat{\Theta}_0 = -\int_0^t \nabla_\theta L(\widehat{\Theta}_{\lfloor \frac{s}{\eta}\rfloor\eta}) \dif s
+ \sqrt{\eta} \int_0^t C(\widehat{\Theta}_{\lfloor\frac{s}{\eta}\rfloor\eta}) \dif W_s \ , \ \widehat{\Theta}_0=\theta_0 \ .
\end{equation}

Here for a real positive number $a>0$ we define $\lfloor a \rfloor=\max\left\{k\in \mathbb{N}_+, k<a\right\}$.
From \eqref{Eq:HatTheta} we see that we have, for $k=0,1,2,...$

\begin{equation}\label{Eq:HatThetaInterpolationSameGLD}
\widehat{\Theta}_{(k+1)\eta}-\widehat{\Theta}_{k\eta}=-\eta \nabla_{\theta}L(\widehat{\Theta}_{k\eta})-\sqrt{\eta}C(\widehat{\Theta}_{k\eta})(W_{(k+1)\eta}-W_{k\eta}) \ .
\end{equation}

Since $\sqrt{\eta}(W_{(k+1)\eta}-W_{k\eta})\sim \mathcal{N}(0, \eta^2 I)$, 
we could let $\eta \Wcal_{k+1}=\sqrt{\eta}(W_{(k+1)\eta}-W_{k\eta})$, where $\Wcal_{k+1}$ is the i.i.d. Gaussian sequence in
\eqref{eq:M-SGD-gaussian}.
From here, we see that
\begin{equation}\label{Eq:EqualityGLDHatTheta}
\widehat{\Theta}_{k\eta}=\theta_k \ ,
\end{equation}
where $\theta_k$ is the solution to \eqref{eq:M-SGD-gaussian}.

We first bound $\widehat{\Theta}_t$ in Eq.~\eqref{Eq:HatTheta} and $\Theta_t$ in Eq.~\eqref{eq:sde}.
Then we could obtain the error estimation of $\theta_k=\widehat{\Theta}_{k\eta}$ and $\Theta_{k\eta}$ by simply set $t=k\eta$.

Since we assumed that $\nabla_\theta \ell_i(\theta)$ is $L$--Lipschitz continuous, we get $|C(\theta_1)-C(\theta_2)|=\dfrac{1}{\sqrt{bN}}\sqrt{\sum\limits_{i=1}^N|\nabla_\theta \ell_i(\theta_1)-\nabla_\theta \ell_i(\theta_2)|^2}\leq \dfrac{1}{\sqrt{bN}}\sqrt{NL^2|\theta_1-\theta_2|^2}\leq L|\theta_1-\theta_2|$ since $b\geq 1$. Thus $C(\theta)$ is also $L$--Lipschitz continuous. 
Take a difference between \eqref{Eq:HatTheta} and \eqref{eq:sde} we get

\begin{equation}\label{Eq:DifferenceHatThetaAndTheta}
\widehat{\Theta}_t-\Theta_t=-\int_0^t [\nabla_\theta L(\Theta_{\lfloor \frac{s}{\eta} \rfloor\eta})-\nabla_\theta L(\Theta_s)] \dif s
+\sqrt{\eta}\int_0^t [C(\widehat{\Theta}_{\lfloor \frac{s}{\eta}\rfloor})-C(\Theta_s)]\dif W_s \ .
\end{equation}

We can estimate
\begin{equation}\label{Eq:LLipschitzDifferenceGradLossVector}
\begin{aligned}
&|\nabla_\theta L(\widehat{\Theta}_{\lfloor\frac{s}{\eta}\rfloor\eta})-\nabla_\theta L(\Theta_s)|^2 \\
\leq & 2|\nabla_\theta L(\widehat{\Theta}_{\lfloor\frac{s}{\eta}\rfloor\eta})-\nabla_\theta L(\Theta_{\lfloor\frac{s}{\eta}\rfloor\eta})|^2
+ 2|\nabla_\theta L(\Theta_{\lfloor\frac{s}{\eta}\rfloor\eta})-\nabla_\theta L(\Theta_s)|^2
\\
\leq & 2L^2|\widehat{\Theta}_{\lfloor\frac{s}{\eta}\rfloor\eta}-\Theta_{\lfloor\frac{s}{\eta}\rfloor\eta}|^2
+2L^2 |\Theta_{\lfloor\frac{s}{\eta}\rfloor\eta}-\Theta_s|^2 \ ,
\end{aligned}
\end{equation}
where we used the inequality $|\nabla_\theta L(\theta_1)-\nabla_\theta L(\theta_2)|\leq \dfrac{1}{N}\sum\limits_{i=1}^N
|\nabla_\theta \ell_i(\theta_1)-\nabla_\theta \ell_i(\theta_2)|\leq L |\theta_1-\theta_2|$. 

Similarly, we estimate
\begin{equation}\label{Eq:LLipschitzDifferenceHalfDiffusionMatrix}
\begin{aligned}
& |C(\widehat{\Theta}_{\lfloor\frac{s}{\eta}\rfloor\eta})-C(\Theta_s)|^2
\\
\leq & 2|C(\widehat{\Theta}_{\lfloor\frac{s}{\eta}\rfloor\eta})-C(\Theta_{\lfloor\frac{s}{\eta}\rfloor\eta})|^2
+ 2|C(\Theta_{\lfloor\frac{s}{\eta}\rfloor\eta})-C(\Theta_s)|^2
\\
\leq & 2L^2|\widehat{\Theta}_{\lfloor\frac{s}{\eta}\rfloor\eta}-\Theta_{\lfloor\frac{s}{\eta}\rfloor\eta}|^2
+2L^2 |\Theta_{\lfloor\frac{s}{\eta}\rfloor\eta}-\Theta_s|^2 \ .
\end{aligned}
\end{equation}

On the other hand, from \eqref{Eq:DifferenceHatThetaAndTheta}, the It\^{o}'s isometry~\cite{oksendal2003stochastic} and Cauchy--Schwarz inequality we have
\begin{equation}\label{Eq:SquareDifferenceHatThetaAndTheta}
\begin{aligned}
& \Ebb|\widehat{\Theta}_t-\Theta_t|^2
\\
\leq & \displaystyle{2\Ebb\left|\int_0^t [\nabla_\theta L(\Theta_{\lfloor \frac{s}{\eta} \rfloor\eta})-\nabla_\theta L(\Theta_s)] \dif s\right|^2
+2\eta\Ebb \left|\int_0^t [C(\widehat{\Theta}_{\lfloor \frac{s}{\eta}\rfloor})-C(\Theta_s)]\dif W_s\right|^2}
\\
\leq & \displaystyle{2\Ebb\left|\int_0^t [\nabla_\theta L(\Theta_{\lfloor \frac{s}{\eta} \rfloor\eta})-\nabla_\theta L(\Theta_s)] \dif s\right|^2
+2\eta \int_0^t \Ebb\left|C(\widehat{\Theta}_{\lfloor \frac{s}{\eta}\rfloor})-C(\Theta_s)\right|^2\dif s}
\\
\leq &
\displaystyle{2\int_0^t \Ebb\left|\nabla_\theta L(\Theta_{\lfloor \frac{s}{\eta} \rfloor\eta})-\nabla_\theta L(\Theta_s)\right|^2\dif s
+2\eta \int_0^t \Ebb\left|C(\widehat{\Theta}_{\lfloor \frac{s}{\eta}\rfloor})-C(\Theta_s)\right|^2\dif s \ .}
\end{aligned}
\end{equation}

Combining \eqref{Eq:LLipschitzDifferenceGradLossVector}, \eqref{Eq:LLipschitzDifferenceHalfDiffusionMatrix} and
\eqref{Eq:SquareDifferenceHatThetaAndTheta} we obtain that
\begin{equation}\label{Eq:SquareDifferenceHatThetaAndThetaGronwall-1}
\begin{aligned}
& \Ebb|\widehat{\Theta}_t-\Theta_t|^2
\\
\leq &
\displaystyle{2\int_0^t \left(2L^2\Ebb |\widehat{\Theta}_{\lfloor\frac{s}{\eta}\rfloor\eta}-\Theta_{\lfloor\frac{s}{\eta}\rfloor\eta}|^2+2L^2\Ebb|\Theta_{\lfloor \frac{s}{\eta} \rfloor\eta}-\Theta_s|^2\right)\dif s}
\\
& \displaystyle{+2\eta \int_0^t \left(2L^2\Ebb|\widehat{\Theta}_{\lfloor\frac{s}{\eta}\rfloor\eta}-\Theta_{\lfloor\frac{s}{\eta}\rfloor\eta}|^2
+2L^2 \Ebb |\Theta_{\lfloor\frac{s}{\eta}\rfloor\eta}-\Theta_s|^2\right)\dif s \ .}
\\
=&\displaystyle{4(1+\eta) L^2\cdot \left(\int_0^t \Ebb|\widehat{\Theta}_{\lfloor\frac{s}{\eta}\rfloor\eta}-\Theta_{\lfloor\frac{s}{\eta}\rfloor\eta}|^2\dif s
+\int_0^t \Ebb |\Theta_{\lfloor\frac{s}{\eta}\rfloor\eta}-\Theta_s|^2 \dif s\right) \ .}
\end{aligned}
\end{equation}

Since we assumed that there is an $M>0$ such that $\max\limits_{i=1,2...,N}(|\nabla_\theta \ell_i(\theta)|)\leq M$, we conclude that $|\nabla_\theta L(\theta)|\leq \dfrac{1}{N}\sum\limits_{i=1}^N |\nabla_\theta \ell_i(\theta)|\leq M$ and $|C(\theta)|\leq \dfrac{1}{\sqrt{bN}}\sqrt{\sum\limits_{i=1}^N |\nabla_\theta \ell_i(\theta)|^2}\leq M$ since $b\geq 1$. 
By \eqref{eq:sde}, the It\^{o}'s isometry~\cite{oksendal2003stochastic}, the Cauchy-Schwarz inequality
and $0\leq s-\lfloor\frac{s}{\eta}\rfloor\eta \leq \eta$ we know that

\begin{equation}\label{Eq:SquareDifferenceThetaDiscretization}
\begin{aligned}
&  \Ebb |\Theta_{\lfloor\frac{s}{\eta}\rfloor\eta}-\Theta_s|^2
\\
= & \displaystyle{\Ebb \left|-\int_{\lfloor\frac{s}{\eta}\rfloor\eta}^s \nabla_{\theta}L(\Theta_u) \dif u
+\sqrt{\eta}\int_{\lfloor\frac{s}{\eta}\rfloor\eta}^s C(\Theta_u)\dif W_u \right|^2}
\\
\leq & \displaystyle{2\Ebb \left|\int_{\lfloor\frac{s}{\eta}\rfloor\eta}^s \nabla_{\theta}L(\Theta_u) \dif u\right|^2
+2\eta\Ebb\left|\int_{\lfloor\frac{s}{\eta}\rfloor\eta}^s C(\Theta_u)\dif W_u \right|^2}
\\
\leq & \displaystyle{2\Ebb \left(\int_{\lfloor\frac{s}{\eta}\rfloor\eta}^s \left|\nabla_{\theta}L(\Theta_u)\right|\dif u\right)^2
+2\eta\int_{\lfloor\frac{s}{\eta}\rfloor\eta}^s \Ebb|C(\Theta_u)|^2 \dif u}
\\
\leq & \displaystyle{2\eta \int_{\lfloor\frac{s}{\eta}\rfloor\eta}^s \Ebb|\nabla_{\theta}L(\Theta_u)|^2 \dif u
+2\eta \int_{\lfloor\frac{s}{\eta}\rfloor\eta}^s \Ebb|C(\Theta_u)|^2 \dif u}
\\
\leq & \displaystyle{2\eta^2 M^2
+2\eta^2 M^2=4\eta^2 M^2 \ .}
\end{aligned}
\end{equation}

Combining \eqref{Eq:SquareDifferenceThetaDiscretization} and \eqref{Eq:SquareDifferenceHatThetaAndThetaGronwall-1}
we obtain
\begin{equation}
\Ebb|\widehat{\Theta}_t-\Theta_t|^2\leq
4(1+\eta)L^2 \cdot \left(\int_0^t \Ebb|\widehat{\Theta}_{\lfloor\frac{s}{\eta}\rfloor\eta}-\Theta_{\lfloor\frac{s}{\eta}\rfloor\eta}|^2\dif s
+4\eta^2 M^2 t\right) \ .
\end{equation}

Set $T>0$ and $m(t)=\max\limits_{0\leq s\leq t}\Ebb |\widehat{\Theta}_s-\Theta_s|^2$, noticing that
$m(\lfloor\frac{s}{\eta}\rfloor \eta)\leq m(s)$ (as $\lfloor\frac{s}{\eta}\rfloor\eta \leq s$), then the above gives for any $0\leq t\leq T$,
\begin{equation}\label{Eq:SquareDifferenceHatThetaAndThetaGronwall-2}
m(t) \leq
4(1+\eta) L^2\cdot \left(\int_0^t m(s)\dif s
+4\eta^2 M^2 T\right) \ .
\end{equation}

By Gronwall's inequality we obtain that for $0\leq t \leq T$,
\begin{equation}
m(t) \leq
16(1+\eta) L^2 \eta^2 M^2 T e^{4(1+\eta) L^2 t}.
\end{equation}
Suppose $0<\eta<1$, then there is a constant $C$ which is independent on $\eta$ s.t.
\begin{equation}\label{Eq:SquareDifferenceHatThetaAndThetaGronwall-Final}
    \Ebb|\widehat{\Theta}_t-\Theta_t|^2 \le m(t) \leq C \eta^2.
\end{equation}
Set $t=k\eta$ in \eqref{Eq:SquareDifferenceHatThetaAndThetaGronwall-Final} and make use of \eqref{Eq:EqualityGLDHatTheta},
we finish the proof.

\end{proof}

\begin{rmk*}
As we have seen in the previous proof, the functions $\nabla_\theta L(\theta)$ and $C(\theta)$ are both $L$--Lipschitz continuous, and thus the SDE \eqref{eq:sde} admits a unique solution (\cite{oksendal2003stochastic}, Section 5.2).
\end{rmk*}

\section{Experiments setups and further results}\label{sec:setups}

The experiments are conducted using GeForce GTX 1080 Ti and \texttt{PyTorch 1.0.0}.

\subsection{FashionMNIST}
\paragraph{Dataset}
\url{https://github.com/zalandoresearch/fashion-mnist}

We randomly choose $1,000$ original test data as our training set, and use the $60,000$ original training data as our test set. Thus we have $1,000$ training data and $60,000$ test data.
We scale the image data to $[0,1]$.

\paragraph{Model}
We use a LeNet alike convolutional network:
\begin{align*}
    \text{input} &\Rightarrow \text{conv1} \Rightarrow \text{max\_pool} \Rightarrow \text{ReLU} \Rightarrow \text{conv2} \Rightarrow \\
    &\text{max\_pool} \Rightarrow \text{ReLU}
    \Rightarrow \text{fc1} \Rightarrow \text{ReLU} \Rightarrow \text{fc2} \Rightarrow \text{output}.
\end{align*}
Both convolutional layers use $5\times 5$ kernels with $10$ channels and no padding. The number of hidden units between fully connected layers are $50$. The total number of parameters of this network are $11,330$.

\paragraph{Optimization}
We use standard (stochastic) gradient descent optimizer. The learning rate is $0.01$.
If not stated otherwise, the batch size of SGD is $50$.

\subsection{SVHN}
\paragraph{Dataset}
\url{http://ufldl.stanford.edu/housenumbers/}

We randomly choose $25,000$ original test data as our training set, and $75,000$ original training data as our test set. Thus we have $25,000$ training data and $75,000$ test data.
We scale the image data to $[0,1]$.

\paragraph{Model}
We use standard VGG-11 without Batch Normalization.

\paragraph{Optimization}
We use standard (stochastic) gradient descent optimizer. The learning rate is $0.05$.
If not stated otherwise, the batch size of SGD is $100$.

\subsection{CIFAR-10}

\paragraph{Dataset}
\url{https://www.cs.toronto.edu/~kriz/cifar.html}

We use standard CIFAR-10 dataset.
We scale the image into $[0,1]$.

\paragraph{Models}
We use two models: VGG-11 without Batch Normalization and standard ResNet-18.

\paragraph{Optimization for VGG-11}
We use momentum (stochastic) gradient descent optimizer. The momentum is $0.9$. The learning rate is $0.01$ decayed by $0.1$ at iteration $40,000$ and $60,000$.
If not stated otherwise, the batch size of SGD is $100$.

\paragraph{Optimization for ResNet-18}
We use momentum (stochastic) gradient descent optimizer. The momentum is $0.9$. The learning rate is $0.1$ decayed by $0.1$ at iteration $40,000$ and $60,000$.
If not stated otherwise, the batch size of SGD is $100$.

For large batch training, we use ghost batch normalization~\cite{hoffer2017}.

Specially, for the experiments to obtain state-of-the-art performance on ResNet-18, we also use standard data augmentation and weight decay $5\times 10^{-4}$.

\subsection{Additional experiments}

\begin{figure}
\centering
\begin{tabular}{cc}
\includegraphics[width=0.45\linewidth]{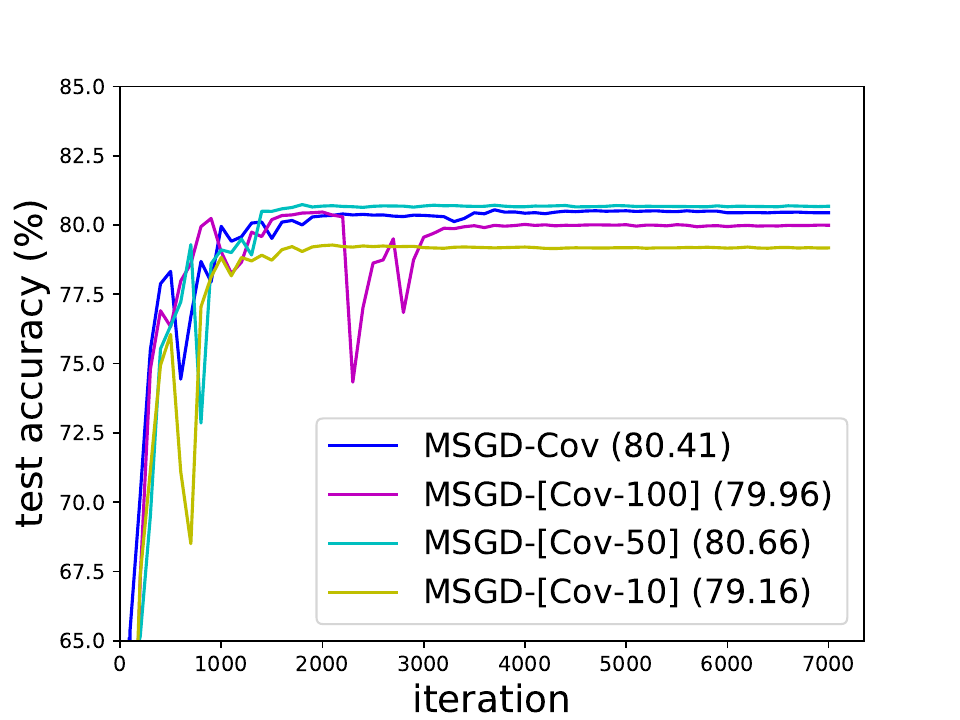} & 
\includegraphics[width=0.45\linewidth]{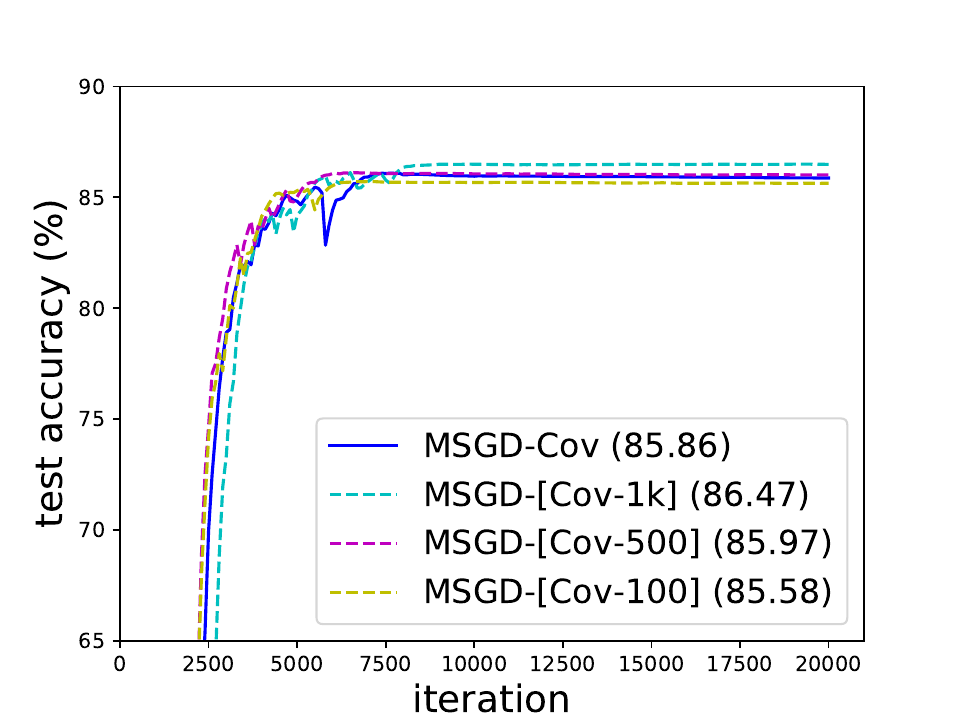}\\
(a) Small FashionMNIST  & (b) SVHN
\end{tabular}
\caption{
The generalization of MSGD.
X-axis: number of iterations; y-axis: test accuracy.
\textbf{(a)}: We randomly draw $1,000$ samples from FashionMNIST as the training set, then train a small convolutional network with them.
\textbf{(b)}: We use $25,000$ samples from SVHN as the training set, then train a VGG-11 without Batch Normalization.
\textbf{MSGD-Cov}: MSGD with Gaussian gradient noise whose covariance is the SGD covariance.
\textbf{MSGD-[Cov-$\mathbf{B}$]}: MSGD-Cov with the SGD covariance estimated using a mini-batch of samples in size $B$.
}
\label{fig:msgd-cov}
\end{figure}

\paragraph{FashionMNIST and SVHN}
Figure~\ref{fig:msgd-cov} shows additional experiments for MSGD-Cov.
We see that indeed for MSGD-Cov, 1) the performance is similar to MSGD-Fisher, and 2) noises from different classes can generalize similarly.

\begin{figure}
\centering
\begin{tabular}{cc}
\includegraphics[width=0.45\linewidth]{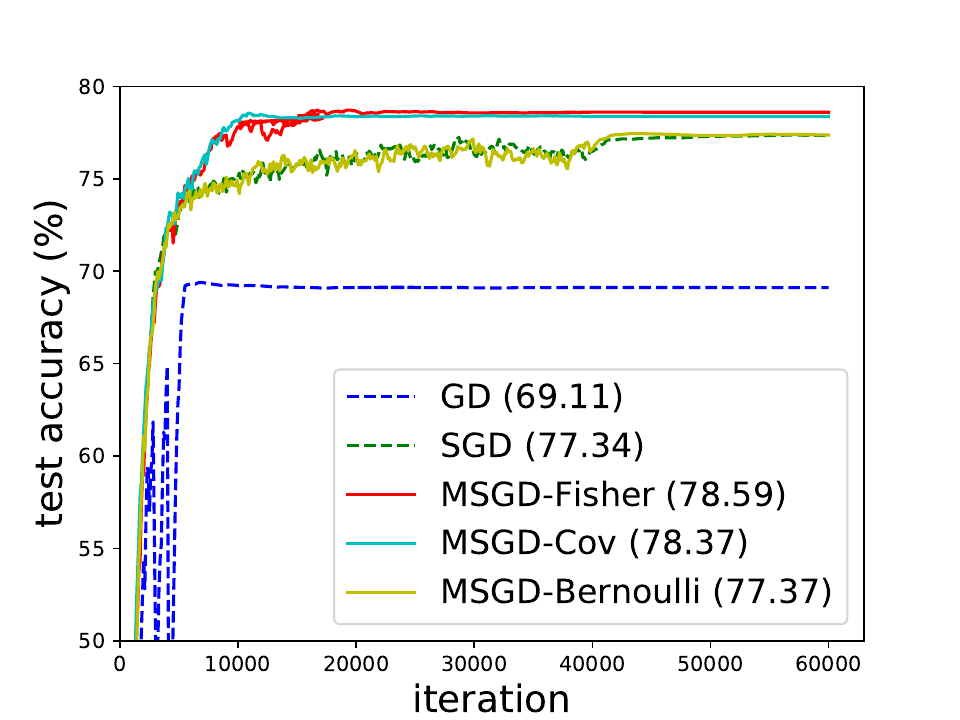} & 
\includegraphics[width=0.45\linewidth]{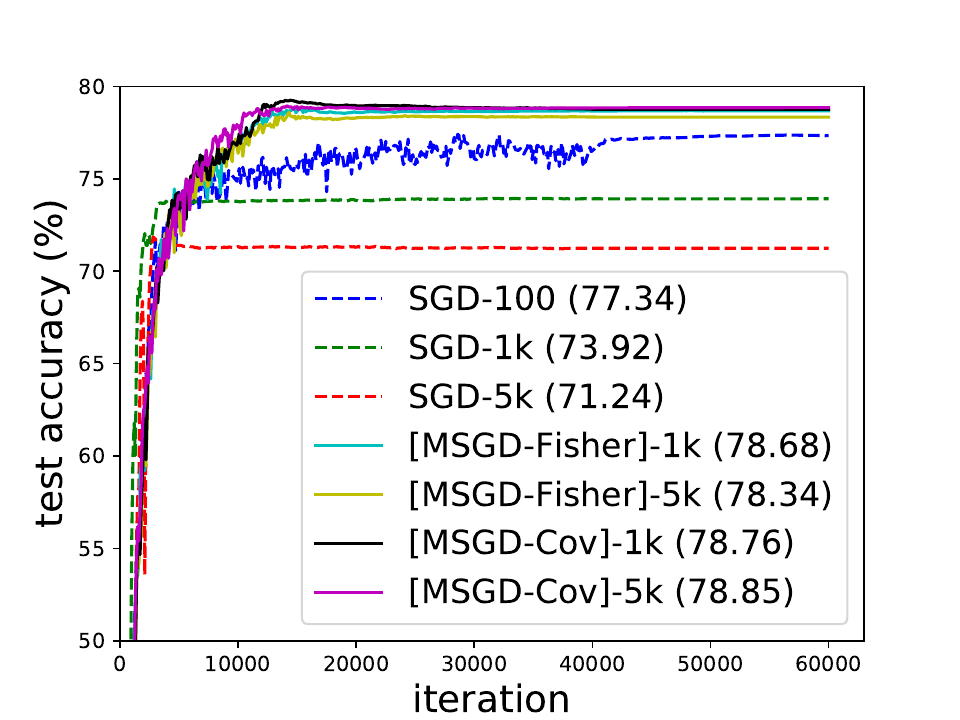}\\
(a) CIFAR-10, VGG-11 & (b) CIFAR-10, VGG-11
\end{tabular}
\caption{
The generalization of MSGD and mini-batch MSGD.
X-axis: number of iterations; y-axis: test accuracy.
\textbf{(a) (b)}: We train a VGG-11 on CIFAR-10 without using Batch Normalization, data augmentation and weight decay.
\textbf{MSGD-Fisher}: MSGD with Gaussian gradient noise whose covariance is the scaled Fisher.
\textbf{MSGD-Cov}: MSGD with Gaussian gradient noise whose covariance is the SGD covariance.
\textbf{MSGD-Bernoulli}: MSGD with Bernoulli sampling noise.
\textbf{SGD-$\mathbf{B}$}: SGD with batch size $B$.
\textbf{[MSGD-Fisher]-$\mathbf{B}$}: mini-batch MSGD with batch size $B$, and an compensatory gradient noise whose covariance is the estimated Fisher.
textbf{[MSGD-Cov]-$\mathbf{B}$}: mini-batch MSGD with batch size $B$, and an compensatory gradient noise whose covariance is the estimated SGD covariance.
}
\label{fig:cifar-vgg}
\end{figure}

\paragraph{VGG-11}
Figure~\ref{fig:cifar-vgg} repeats our experiments in main text on VGG-11. The results are consistent with our main conclusions.

\paragraph{CIFAR-100}
Table~\ref{tab:cifar-100} show addtional result for CIFAR-100 on ResNet-18. The setups follow Figure~\ref{fig:batch-msgd} (c), except that the dataset is CIFAR-100 instead of CIFAR-10.

\begin{table}
\caption{Additioal experimetns for CIFAR-100 on ResNet-18}
\label{tab:cifar-100}
\centering
\begin{tabular}{c|c}
\hline
Algorithm & Test Accuracy \\
\hline\hline
SGD-500 & $76.38\%$ \\
SGD-2k & $72.78\%$ \\
{[MSGD-Fisher]-2k} & $76.83\%$ \\
SGD-5k & $59.16\%$ \\
{[MSGD-Fisher]-5k} & $76.46\%$ \\
\hline
\end{tabular}
\end{table}

\end{document}